\documentclass[10pt,journal,cspaper,compsoc]{IEEEtran}

\ifCLASSOPTIONcompsoc
\usepackage[nocompress]{cite}
\else
\fi

\usepackage[strict]{changepage}

\usepackage{multirow}
\usepackage{amssymb}
\usepackage{booktabs}
\usepackage{xcolor}
\usepackage[english]{babel}
\usepackage[pdftex]{graphicx}
\usepackage{subcaption}
\usepackage{amsmath}
\usepackage{mathpazo}
\usepackage{cancel}
\usepackage{mathabx}

\DeclareMathOperator{\Tr}{Tr}

\usepackage{url}
\usepackage{hyperref}

\definecolor{mygray}{rgb}{.0,.0,.8}
\definecolor{mygreen}{rgb}{.0,.8,.0}
\definecolor{myblack}{rgb}{0,0,0}
\newcommand{\added}[1]{\textcolor{myblack}{#1}}

\newsavebox{\fourfigurebox}

\begin{document}
\title{The Shape of Learning Curves: a Review}
\author{Tom~Viering,%
        ~Marco~Loog%
\IEEEcompsocitemizethanks{\IEEEcompsocthanksitem T.~Viering is with Delft University of Technology, the Netherlands.\protect\\
E-mail: t.j.viering@tudelft.nl
\IEEEcompsocthanksitem M.~Loog is with Delft University of Technology, the Netherlands, and with the University of Copenhagen, Denmark.}%
\thanks{Manuscript received 2021}}

\markboth{IEEE Transactions on Pattern Analysis and Machine Intelligence}%
{The Shape of Learning Curves: a Review}
\IEEEtitleabstractindextext{%

\begin{abstract}
Learning curves provide insight into the dependence of a learner’s generalization performance on the training set size. This important tool can be used for model selection, to predict the effect of more training data, and to reduce the computational complexity of model training and hyperparameter tuning. This review recounts the origins of the term, provides a formal definition of the learning curve, and briefly covers basics such as its estimation. Our main contribution is a comprehensive overview of the literature regarding the shape of learning curves. We discuss empirical and theoretical evidence that supports well-behaved curves that often have the shape of a power law or an exponential. We consider the learning curves of Gaussian processes, the complex shapes they can display, and the factors influencing them. We draw specific attention to examples of learning curves that are ill-behaved, showing worse learning performance with more training data. To wrap up, we point out various open problems that warrant deeper empirical and theoretical investigation. All in all, our review underscores that learning curves are surprisingly diverse and no universal model can be identified.
\end{abstract}
\begin{IEEEkeywords}
Learning curve, training set size, supervised learning, classification, regression.
\end{IEEEkeywords}}
\maketitle
\IEEEdisplaynontitleabstractindextext
\IEEEpeerreviewmaketitle
\IEEEraisesectionheading{\section{Introduction}\label{sec:introduction}}
\bstctlcite{BSTcontrol}

\begin{figure}[tb]
	\centering
	\includegraphics[width=0.84\linewidth]{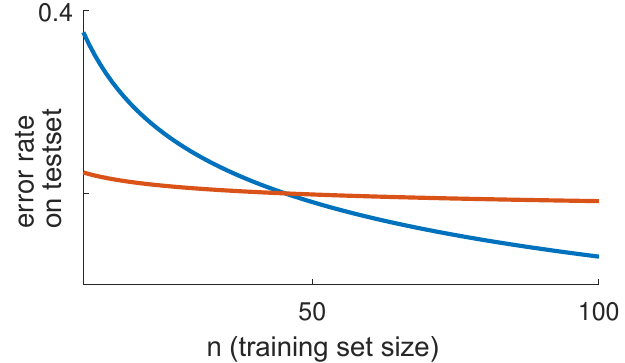}
	\caption{\added{Two crossing learning curves.} Red starts at a lower error, while blue reaches a lower error rate given enough data. One number summaries cannot characterize such behaviors. For more details see Section \ref{sect:aulc} and \ref{sect:cross}.}
	\label{fig:smallcrossing}
\end{figure}

\IEEEPARstart{T}{he} more often we are confronted with a particular problem to solve, the better we typically get at it.  The same goes for machines.  A \emph{learning curve} is an important, graphical representation that can provide insight into such learning behavior by plotting generalization performance against the number of training examples. \added{Two example curves are shown in Figure \ref{fig:smallcrossing}.}

We review learning curves in the context of standard supervised learning problems such as classification and regression.  The primary focus is on the shapes that learning curves can take on. We make a distinction between well-behaved learning curves that show improved performance with more data and ill-behaved learning curves that, perhaps surprisingly, do not. We discuss theoretical and empirical evidence in favor of different shapes, underlying assumptions made, how knowledge about those shapes can be exploited, and further results of interest.  In addition, we provide the necessary background to interpret and use learning curves as well as a comprehensive overview of the important research directions.

\subsection{Outline}

The next section starts off with a definition of learning curves and discusses how to estimate them in practice.  It also briefly considers  so-called \emph{feature curves}, which offer a complementary view.  Section \ref{sect:use} covers the use of learning curves, such as the insight into model selection they can give us, and how they are employed, for instance, in meta-learning and reducing the cost of labeling or computation.  Section \ref{sec:well_behaved} considers evidence supporting well-behaved learning curves: curves that generally show improved performance with more training data.  We review the parametric models that have been studied empirically and cover the theoretical findings in favor of some of these.  Many of the more theoretical results in the literature have been derived particularly for Gaussian process regression as its learning curve is more readily analyzed analytically. Section \ref{sec:GP} is primarily devoted to those specific results. Section \ref{sect:ill} then follows with an overview of important cases of learning curves that do not behave well and considers possible causes. We believe that especially this section shows that our understanding of the behavior of learners is more limited than one might expect. Section \ref{sect:disc} provides an extensive discussion. It also concludes our review.  The remainder of the current section goes into the origins and meanings of the term ``learning curve'' and its synonyms.  

\subsection{Learning Curve Origins and Meanings}\label{sect:term}

With his 1885 book \emph{{\"U}ber das Ged{\"a}chtnis} \cite{ebbinghaus1885gedachtnis},  Ebbinghaus is generally considered to be the first to employ and qualitatively describe learning curves.  These curves report on the number of repetitions it takes a human to perfectly memorize an increasing number of meaningless syllables.  Such learning curves have found widespread application for predicting  human productivity, and are the focus of previous surveys on learning curves  \cite{yelle1979learning,anzanello2011learning}. 

While Ebbinghaus is the originator of the learning curve concept, it should be noted that the type of curves from \cite{ebbinghaus1885gedachtnis,yelle1979learning,anzanello2011learning} are importantly different from the curves central to this review. 
In the machine learning setting, we typically care about the generalization performance, i.e., the learner's performance on \emph{new and unseen} data. Instead, Ebbinghaus's subject goal is to recite exactly that string of syllables that has been provided to him. Similarly, studies of human productivity focus on the speed and cost of repetitions of the same task. This means in a way that the performance on the training set is considered.  This measure is also called the resubstitution or apparent error in classification \cite{jain2000statistical,loog2018supervised,schiavo2000ten}.
Indeed, as most often is the case for the training error as well, memorization performance gets worse as the amount of training data increases, i.e., it is more difficult to memorize an increasing number of syllables.

The term learning curve considered in this review is different from the curve that displays the training error---or the value of any objective function---as a function of the number of epochs or iterations used for optimization. \added{Especially in the neural network literature, this is what the learning curve often signifies\footnote{\added{It should be noted that such plots have to be interpreted with care, as the number of optimization steps or similar quantities can dependent on the training set size, the batch size, etc.  Precise specifications are key therefore.}} \cite{domhan2015speeding,Xincollection,sammut2011encyclopedia}.} What it has in common with those of Ebbinghaus is that the performance is plotted against the number of times that (part of) the data has been revisited, which corresponds directly to the number of repetitions in \cite{ebbinghaus1885gedachtnis}.  These curves, used to monitor the optimality of a learner in the training phase, are also referred to as \emph{training curves} and this terminology can be traced back to \cite{osborn1975}.
We use training curve exclusively to refer to these curves that visualize performance during training.
Many researchers and practitioners, however, use the term learning curve instead of training curve  \cite{Xincollection}, which, at times, may lead to confusion. 

In the machine learning literature synonyms for learning curve are \emph{error curve}, \emph{experience curve}, \emph{improvement curve} and \emph{generalization curve} \cite{Xincollection, atlas1990training,sompolinsky1990learning}. 
Improvement curve can be traced back to a 1897 study, on learning the telegraphic language \cite{bryan1897studies}. 
Generalization curve was first used in machine learning in 1990 \cite{atlas1990training,sompolinsky1990learning}.  Decades earlier, the term was already used to indicate the plot of the intensity of an animal's response against stimulus size \cite{spence1937differential}.
Learning curve or, rather, its German equivalent was not used as an actual term in the original work of Ebbinghaus \cite{ebbinghaus1885gedachtnis}. The English variant seems to appear 18 years later, in 1903 \cite{swift1903studies}. \emph{Lehrkurve} follows a year after \cite{lipmann1904einfluss}.

We traced back the first mention of learning curve in connection to learning machines to a discussion in an 1957 issue of \emph{Philosophy} \cite{ritchie1957thinking}.  A year later, Rosenblatt, in his famous 1958 work \cite{rosenblatt1958perceptron}, uses learning curves in the analysis of his perceptron. Following this review's terminology, he uses this term to refer to a training curve.  %
Foley \cite{Foley1972} was possibly the first to use a learning curve, as it is defined in this review, in an experimental setting such as is common nowadays. The theoretical study of learning curves for supervised learners dates back at least to 1965 \cite{Cover1965}. %

\section{Definition, Estimation, Feature Curves}\label{sect:prelim}

This section makes the notion of a learning curve more precise and describes how learning curves can be estimated from data. We give some recommendations when it comes to plotting learning curves and summarizing them. Finally, feature curves offer a view on learners that is complementary to that of learning curves. These and combined learning-feature curves are covered at the end of this section.

\subsection{Definition of Learning Curves} \label{sub:def}

Let $S_n$ indicate a training set of size $n$, which acts as input to some learning algorithm $A$. In standard classification and regression, $S_n$ consists of $(x,y)$ pairs, where $x \in \mathbb{R}^d$ is the $d$-dimensional input vector (i.e., the features, measurements, or covariates) and $y$ is the corresponding output (e.g. a class label or regression target).  $\mathcal{X}$ denotes the input space and $\mathcal{Y}$ the output space. The $(x,y)$ pairs of the training set are i.i.d. samples of an unknown probability distribution $P_{XY}$ over $\mathcal{X} \times \mathcal{Y}$. Predictors $h$ come from the hypothesis class $\mathcal{H}$, which contains all models that can be returned by the learner $A$. An example of a hypothesis class is the set of all linear models $\{h:x\mapsto a^Tx + b|a\in\mathbb{R}^d,b\in\mathbb{R}\}$.

When $h$ is evaluated on a sample $x$, its prediction for the corresponding $y$ is given by $\hat{y} = h(x)$. The performance of a particular hypothesis $h$ is measured by a loss function $L$ that compares $y$ to $\hat{y}$. Examples are the squared loss for regression, where $\mathcal{Y} \subset \mathbb{R}$ and $L_\text{sq}(y,\hat{y}) = (y-\hat{y})^2$ and the zero-one loss for (binary) classification $L_\text{01}(y,\hat{y})= \tfrac{1}{2}(1-y\hat{y})$ when $\mathcal{Y}=\{-1,+1\}$.

The typical goal is that our predictor performs well on average on all new and unseen observations. Ideally, this is measured by the expected loss or risk $R$ over the true distribution $P_{XY}$:
\begin{equation}
    R(h) = \int L(y,h(x)) dP(x,y). \label{eq_risk}
\end{equation}
Here, as in most that follows, we omit the subscript $XY$.

Now, an \emph{individual} learning curve considers a single training set $S_n$ for every $n$ and calculates its corresponding risk $R(A(S_n))$ as a function of $n$. \added{Note that training sets can be partially ordered, meaning $S_{n-1} \subset S_n$.} However, a single $S_n$ may deviate significantly from the expected behavior. Therefore, we are often interested in an averaging over many different sets $S_n$, and ideally the expectation 
\begin{equation}
	\bar{R}_n(A) = \underset{S_n \sim P^n}{\mathbb{E}} R(A(S_n)). \label{eq_exp_lc}
\end{equation}
The plot of $\bar{R}_n(A)$ against the training set size $n$ gives us the (expected) learning curve. From this point onward, when we talk about \emph{the} learning curve, this is what is meant. 

The preceding learning curve is defined for a single problem $P$. Sometimes we wish to study how a model performs over a range of problems or, more generally, a full distribution $\mathcal{P}$ over problems.  The learning curve that considers such averaged performance is referred to as the problem-average (PA) learning curve:
\begin{equation}\label{eq:pa}
\bar{R}^\text{PA}_n(A) = \underset{P \sim \mathcal{P}}{\mathbb{E}}  \bar{R}_n(A).
\end{equation}
The general term problem-average was coined in \cite{Duda2012}. PA learning curves make sense for Bayesian approaches in particular, where an assumed prior over possible problems often arises naturally. As GP's are Bayesian, their PA curve is frequently studied, see Section \ref{sec:GP}. The risk integrated over the prior, in the Bayesian literature, is also called the Bayes risk, integrated risk, or preposterior risk \cite[page~195]{Murphy2012}.  The term preposterior signifies that, in principle, we can determine this quantity without observing any data. 

In semi-supervised learning \cite{chapelle2010semi} and active learning \cite{settles2009active}, it can be of additional interest to study the learning behavior as a function of the number of \emph{unlabeled} and \emph{actively selected} samples, respectively. %

\subsection{Estimating Learning Curves} \label{sub_est_lc}

In practice, we merely have a finite sample from $P$ and we cannot measure $R(h)$ or consider all possible training sets sampled from $P$.  We can only get an estimate of the learning curve. Popular approaches are to use a hold-out dataset or $k$-fold cross validation for this \cite{Mukherjee2003, Figueroa2012, Richter2019, Ng2002} as also apparent from the Weka documentation \cite{weka} and Scikit-learn  implementation \cite{scikit}. %
Using cross validation, $k$ folds are generated from the dataset.  For each split, a training set and a test set are formed. The size of the training set $S_n$ is varied over a range of values by removing samples from the originally formed training set. For each size the learning algorithm is trained and the performance is measured on the test fold. The process is repeated for all $k$ folds, leading to $k$ individual learning curves. The final estimate is their average. 
The variance of the estimated curve can be reduced by carrying out the $k$-fold cross validation multiple times \cite{kim2009estimating}. This is done, for instance, in \cite{Hess2010, Mukherjee2003, Figueroa2012, Richter2019}. %

Using cross validation to estimate the learning curve has some drawbacks. For one, when making the training set smaller, not using the discarded samples for testing seems wasteful. Also note that the training fold size limits the range of the estimated learning curve, especially if $k$ is small.  Directly taking a random training set of the preferred size and leaving the remainder as a test set can be a good alternative. This can be repeated to come to an averaged learning curve. This recipe---employed, for instance in \cite{prtools,Hess2010}---allows for easy use of any range of $n$, leaving no sample unused. Note that the test risks are not independent for this approach. 
\added{Alternatively, drawing samples from the data with replacement can also be used to come to to variable sized training sets (e.g. similar to bootstrapping \cite{efron1983estimating,jain1987bootstrap}).} \added{For classification, often a learning curve is made with stratified training sets to reduce the variance further.}

\added{A learner $A$ often depends on hyperparameters. Tuning once and fixing them to create the learning curve will result in  suboptimal curves, since the optimal hyperparameter values can strongly depend on $n$. Therefore, ideally, the learner should internally use cross validation (or the like) to tune hyperparameters for each training set it receives.}

An altogether different way to learning curve estimation is to assume an underlying parametric model for the learning curve and fit this to the learning curves estimates obtained via approaches described previously. The approach is not widespread under practitioners, but is largely confined to the research work that studies and exploits the general shape of learning curves (see Subsection \ref{sect:param}).

Finally, note that all of the foregoing pertains to PA learning curves as well.  In that setting, we may occasionally be able to exploit the special structure of the assumed problem prior.  This is the case, for instance, with Gaussian process regression, where problem averages can sometimes be computed with no additional cost (Section \ref{sec:GP}). 

\subsection{\added{Plotting Considerations}}\label{sect:plot}
When plotting the learning curve, it can be useful to consider logarithmic axes. Plotting $n$ linearly may mask small but non-trivial gains \cite{Perlich2003}. Also from a computational standpoint it often makes sense to have $n$ traverse a logarithmic scale \cite{Provost1999} (see also Subsection \ref{sec:extrapolate}). A log-log or semi-log plot can be useful if we expect the learning curve to display power-law or exponential behavior (Section \ref{sec:well_behaved}), as the curve becomes straight in that case. In such a plot, it can also be easier to discern small deviation from such behavior.  Finally, it is common to use error bars to indicate the standard deviation over the folds to give an estimate of the variability of the individual curves.

\subsection{Summarizing Learning Curves}\label{sect:aulc}

It may be useful at times, to summarize learning curves into a single number.  A popular metric to that end is the area under the learning curve (AULC) \cite{perez1995using,mazzoni2004active,settles2008analysis} (see \cite{ghiselli1937comparison} for early use). To compute this metric, one needs to settle at a number of sample sizes. One then averages the performance at all those sample sizes to get to the area of the learning curve. %
The AULC thus makes the curious assumption that all sample sizes are equally likely.

Important information can get lost when summarizing.  The measure is, for instance, not able to distinguish between two methods whose learning curves cross (Figure \ref{fig:smallcrossing}), i.e.,  where the one method is better in the small sample regime, while the other is in the large sample setting. Others have proposed to report the asymptotic value of the learning curve and the number of samples to reach it \cite{Langley1988} or the exponent of the power-law fit \cite{bertoldi2012evaluating}.  

Depending on the application at hand, particularly in view of the large diversity in learning curve shapes, these summaries are likely inadequate. Recently, \cite{hoiem2021learning} suggested to summarize using \emph{all} fit parameters, which should suffice for most applications. However, we want to emphasize one should try multiple parametric models and report the parameters of the best fit \emph{and} the fit quality (e.g. MSE).

\subsection{\added{Fitting Considerations}} \label{sec:fitting}

Popular parametric forms for fitting learning curves are given in Table \ref{table:curve_shapes} and their performance is discussed in Section \ref{sect:param}. Here, we make a few general remarks.

Some works \cite{Cortes1994,Frey1999,Singh2005,Last2007} seem to perform simple least squares fitting on log-values in order to fit power laws or exponentials. If, however, one would like to find the optimal parameters in the original space in terms of the mean squared error, non-linear curve fitting is necessary (for example using Levenberg–Marquardt \cite{Gu2001}).  Then again, assuming Gaussian errors in the original space may be questionable, since the loss is typically non-negative. Therefore, confidence intervals, hypothesis tests, and $p$-values should be interpreted with care. 

For many problems, one should consider a model that allows for nonzero asymptotic error (like POW3 and EXP3 in Table \ref{table:curve_shapes}). Also, often the goal is to interpolate or extrapolate to previously \emph{unseen} training set sizes. This is a generalization task and we have to deal with the problems that this may entail, such as overfitting to learning curve data \cite{Gu2001,Kolachina2012}. Thus, learning curve data should also be split in train and test sets for a fair evaluation. 

\subsection{Feature Curves and Complexity}\label{sect:fc}

\begin{figure*}[tbh]
	\centering
	\begin{subfigure}{1\textwidth} 
        \refstepcounter{subfigure}\label{four:a}
        \refstepcounter{subfigure}\label{four:b}
        \refstepcounter{subfigure}\label{four:c}
        \refstepcounter{subfigure}\label{four:d}
    \end{subfigure}%
	\includegraphics[width=0.84\textwidth]{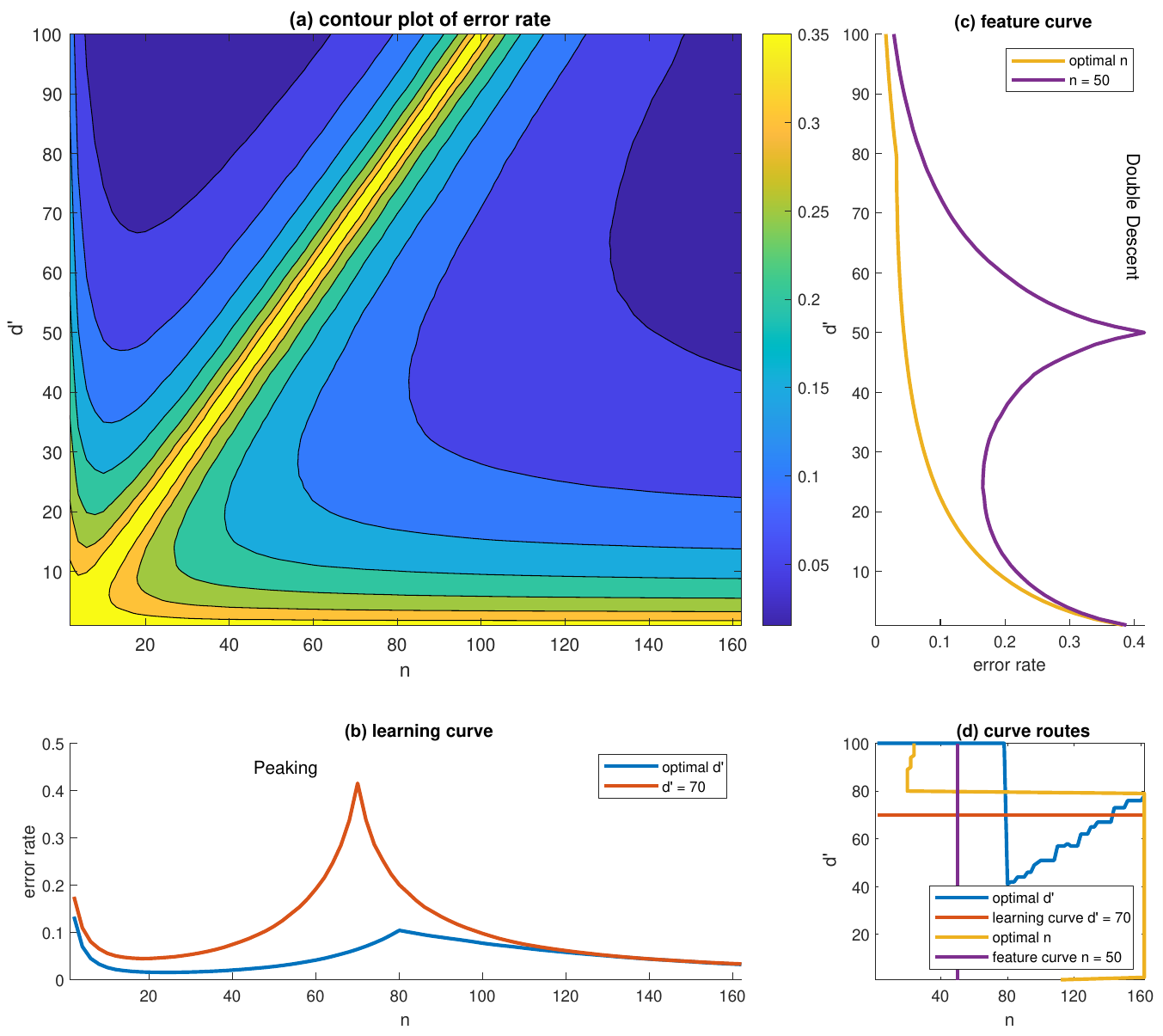}
    
	\caption{(a) Image of the error for varying sample size $n$ and dimensionality $d'$, for the pseudo-Fisher learning algorithm (without intercept) on a toy dataset with two Gaussian classes having identity covariance matrices. Their means are a distance of 6 apart in 100 dimensions, with every dimension adding a same amount to the overall distance. (b) By fixing $d'$ and varying $n$, i.e., taking a horizontal section, we obtain a learning curve (red). We also show the curve where $d'$ is chosen optimally for each $n$ (blue). (c) This plot is rotated by 90 degrees. By fixing $n$ and varying $d'$, i.e., a vertical section, we obtain a feature curve (purple). We also show the curve where the optimal $n$ is chosen for each $d'$ (yellow). (d) Here we show the paths taken through the image to obtain the curves. The learning curve and feature curves are the straight lines, while the curves that optimize $n$ or $d'$ take other paths. The largest $n$ and $d$ are not always optimal. \added{We note that Section \ref{sect:ill} covers learning curves that do not necessarily improve with more data.  Section \ref{sub:peaking} discusses peaking (displayed here).}}
	\label{fig:surfaceplot}
\end{figure*}

The word feature refers to the $d$ measurements that constitutes an input vector $x$. A feature curve is obtained by plotting the performance of a machine learning algorithm $A$ against the varying number of measurements it is trained on \cite{hughes1968mean,Jain1982}.  To be a bit more specific, let $\varsigma_{d'}$ be a procedure that selects $d'$ of the original $d$ features, hence reducing the dimensionality of the data to $d'$. A feature curve is then obtained by plotting $\bar{R}(A(\varsigma_{d'}(S_n)))$ versus $d'$, while $n$ is now the quantity that is fixed.  As such, it gives a view complementary to the learning curve.

The selection of $d'$ features as carried out by means of $\varsigma_{d'}$, can be performed in various ways.  Sometimes features have some sort of inherent ordering.  %
In other cases PCA or feature selection can provide such ordering.
When no ordering can be assumed, $\varsigma_{d'}$ samples $d'$ random features from the data---possibly even with replacement.  In this scenario, it is sensible to construct a large number of different feature curves, based on different random subsets, and report their average as the final curve.

Typically, an increase in the number of input dimensions means that the complexity of the learner also increases.  As such it can, more generally, be of interest to plot performance against any actual, approximate, or substitute measure of the complexity. Instead of changing the dimensionality, changing parameters of the learner, such as the smoothness of a kernel or the amount of filters in a CNN, can also be used to vary the complexity to obtain similar curves \cite{raudys1991small,duin1997experiments,torgo1997kernel,nakkiran2019deep,hughes1968mean}.  Such curves are sometimes called \emph{complexity curves} \cite{torgo1997kernel}, \emph{parameter curves} \cite{duin2005stat} or \emph{generalization curves} \cite{chen2020multiple}.

One of the better known phenomena of feature curves is the so-called peaking phenomenon (also the peak effect, peaking or Hughes phenomenon \cite{van1978peaking, jain198239, raudys1980dimensionality, jain2000statistical}). The peaking phenomenon of feature curves is related to the curse of dimensionality and illustrates that adding features may actually degrade the performance of a classifier, leading to the classical U-shaped feature curve.  %
Behavior more complex than the simple U-shape has been observed as well \cite{vallet1989linear,Duin2000,zollanvari2019theoretical} and has recently been referred to as double descent \cite{Belkin2019} (see Figure \ref{four:a}). This is closely related to peaking of (ill-behaving) learning curves (Subsection \ref{sub:peaking}). 

\subsection{Combined Feature and Learning Curves}\label{sect:comb}

Generally, the performance of a learner $A$ is not influenced independently by the the number of training samples $n$ and the number of features $d$. In fact, several theoretical works suggest that the fraction $\alpha = \tfrac{d}{n}$ is essential (see, e.g. \cite{opper1990ability,watkin1993statistical,engel2001statistical}).
Because of the feature-sample interaction, it can be insightful to plot multiple learning curves for a variable number of input dimensions or multiple feature curves for different training set sizes.  Another option is to make a 3D plot---e.g. a surface plot---or a 2D image of the performance against both $n$ and $d$ directly.  Instead of the number of features any other complexity measure can be used.

Figure \ref{four:a} shows such a plot for pseudo-Fisher's linear discriminant (PFLD; see Section \ref{sub:peaking}) when varying both $n$ and $d'$. Taking a section of this surface, we obtain either a learning curve in Figure \ref{four:b} (horizontal, fixed $d$) or a feature curve in Figure \ref{four:c} (vertical, fixed $n$). Figure \ref{four:a} gives a more complete view of the interaction between $n$ and $d$.  In Figure \ref{four:d} we can see that the optimal $d'$ depends on $n$. Likewise, for this model, there is an optimal $n$ for each $d'$, i.e., the largest possible value of $n$ is not necessarily the best. %

Duin \cite{Duin2000} is possibly the first to include such 3D plot, though already since the work of Hughes \cite{hughes1968mean}, figures that combine multiple learning or feature curves have been used, see for example \cite{jain1978optimal,duin1978accuracy}, and for combinations with complexity curves see   \cite{skurichina1996stabilizing,duin1997experiments,robert1976choice,raudys1991small}. More recently, \cite{nakkiran2019deep,Rosenfeld2020} gives 2D images of the performance of deep neural networks as a function of both model and training set size. 

\section{General Practical Usage}\label{sect:use}

The study of learning curves has both practical and research/theoretical value.  While we do not necessarily aim to make a very strict separation between the two, more emphasis is put on the latter further on. This section focuses on part of the former and covers the current, most important uses of learning curves when it comes to applications, i.e., model selection and extrapolation to reduce data collection and computational costs.  %
\added{For a valuable, complementary review that delves specifically into how learning curves are deployed for decision making, we refer to \cite{mohr2022learning}.}

\subsection{Better Model Selection and Crossing Curves}\label{sect:cross}

Machine learning as a field has shifted more and more to benchmarking learning algorithms, e.g., in the last 20 years, more than 2000 benchmark datasets have been created (see \cite{paperswithcode} for an overview). These benchmarks are often set up as competitions \cite{sculley2018winner} and investigate which algorithms are better or which novel procedure outperforms existing ones \cite{Perlich2003}.  Typically, a single number, summarizing performance, is used as evaluation measure.

A recent meta-analysis indicates that the most popular measures are accuracy, the F-measure, and precision \cite{blagec2020critical}. 
An essential issue these metrics ignore is that sample size can have a large influence on the relative ranking of different learning algorithms. In a plot of learning curves this would be visible as a crossing of the different curves (see Figure \ref{fig:smallcrossing}). In that light, it is beneficial if benchmarks consider multiple sample sizes, to get a better picture of the strengths and weaknesses of the approaches. The learning curve provides a concise picture of this sample size dependent behavior.

Crossing curves have also been referred to as the scissor effect and have been investigated since the 1970s \cite{raudys1970problems,kanal1971dimensionality,duin1978accuracy} (see also \cite{Raudys1998}). 
Contrary to such evidence, there are papers that suggest that learning curves do not cross \cite{kohavi1996scaling,bornschein2020small}. \added{\cite{Perlich2003} calls into question the claim of \cite{kohavi1996scaling} as the datasets may be too small to find the crossing point.}
The latter claim by \cite{bornschein2020small} is specific to deep learning, where, perhaps, exceptions may occur that are currently not understood.

Perhaps the most convincing evidence for crossing curves is given in \cite{Perlich2003}. The paper compares logistic regression and decision trees on 36 datasets. %
In 15 of the 36 cases the learning curves cross. This may not always be apparent, however, as large sample sizes may be needed to find the crossing point.  In the paper, the complex model (decision tree) is better for large sample sizes, while the simple model (logistic regression) is better for small ones.
Similarly, Strang et al.\ \cite{strang2018don} performed a large-scale meta-learning study on 294 datasets, comparing linear versus nonlinear models, and found evidence that non-linear methods are better when datasets are large. Ng and Jordan \cite{Ng2002} found, when comparing naive Bayes to logistic regression, that in 7 out of 15 datasets considered the learning curves crossed. \cite{Cortes1994,Morch1997,shavlik1991symbolic,domingos1997optimality,harris1997sample} provide further evidence.

Also using learning curves, \cite{Perlich2003} finds that, besides sample size, separability of the problem can be an indicator of which algorithm will dominate the other in terms of the learning curve. Beyond that, the learning curve, when plotted together with the training error of the algorithm can be used to detect whether a learner is overfitting \cite{jain2000statistical,duin2005stat,duin2016pattern,loog2018supervised}. Besides sample size, dimensionality seems also an important factor to determine whether linear or non-linear methods will dominate \cite{strang2018don}. To that end, learning curves combined with feature curves may offer further insights. %

\subsection{Extrapolation to Reduce Data Collection Costs}\label{sec:extrapolate}

When collecting data is time-consuming, difficult, or otherwise expensive the possibility to accurately extrapolate a learner's learning curve can be useful. Extrapolations (typically base on some parametric learning curve model, see Subsection \ref{sect:param}) give an impression beforehand of how many examples to collect to come to a specific performance and allows one to judge when data collection can be stopped \cite{Frey1999}.
Examples of such practice can, for instance, be found in machine translation \cite{Kolachina2012} and medical applications \cite{Mukherjee2003, Hess2010, Figueroa2012}.   Last \cite{Last2007} quantifies potential savings assuming a fixed cost per collected sample and per generalization error. Extrapolating the learning curve using some labeled data, the point at which it is not worth anymore to label more data can be determined and data collection stopped.

Determining a minimal sample size is called sample size determination. For usual statistical procedures this is done through what is called a power calculation \cite{jones2003introduction}. %
For classifiers, sample size determination using a power calculation is unfeasible according to \cite{Mukherjee2003, Figueroa2012}. John and Langley \cite{John1996} illustrate that a power calculations that ignores the machine learning model indeed fails to accurately predict the minimal sample size.

Sample size determination can be combined with meta-learning, which uses experience on previous datasets to inform decisions on new datasets. To that end, \cite{Leite2004} builds a small learning curve on a new and unseen dataset and compares it to a database of previously collected learning curves to determine the minimum sample size.

\subsection{Speeding Up Training and Tuning}\label{sec:extrapolate_comp}

Learning curves can be used to reduce computation time and memory with regards to training models, model selection and hyperparameter tuning. %

To speed up training, so-called progressive sampling \cite{Provost1999} uses a learning curve to determine if less training data can reach adequate performance. If the slope of the curve becomes too flat, learning is stopped, making training potentially much faster. It is recommended to use a geometric series for $n$ to reduce computational complexity. 

Several variations on progressive sampling exist. John and Langley \cite{John1996} proposes the notion of \emph{probably close enough} where a power-law fit is used to determine if the learner is so-called epsilon-close to the asymptotic performance. \cite{Meek2002} gives a rigorous decision theoretic treatment of the topic. By assigning costs to computation times and performances, they estimate what should be done to minimize the expected costs. %
Progressive sampling also has been adapted to the setting of active learning \cite{Tomanek2008}. \cite{Leite2004} combines meta-learning with progressive sampling to obtain a further speedup.  

To speed up model selection, \cite{Leite2005} compares initial learning curves to a database of learning curves to predict which of two classifiers will perform best on a new dataset. This can be used to avoid costly evaluations using cross validation. Leite  and  Brazdil \cite{Leite2007} propose an iterative process that predicts the required sample sizes, builds learning curves, and updates the performance estimates in order to compare two classifiers. Rijn et al. \cite{VanRijn2015} extend the technique to rank many machine learning models according to their predicted performance, tuning their approach to come to an acceptable answer in as little time as possible.  \added{\cite{mohr2021fast} does not resort to meta-learning, and instead uses partial learning curves to speed up model selection. This simpler approach can also already lead to significantly improved run times.} 

With regards to hyperparameter tuning, already in 1994 Cortes et al.\ \cite{Cortes1994} devised an extrapolation scheme for learning curves, based on the fitting of power laws, to determine if it is worth to fully train a neural network. In the deep learning era, this has received renewed attention. \cite{Hestness2017} extrapolates the learning curve to optimize hyperparameters. \cite{hoiem2021learning} takes this a step further and actually optimize several design choices, such as data augmentation. %
One obstacle for such applications is that it remains unclear when the learning curve has which shape.

\section{Well-Behaved Learning Curves} \label{sec:well_behaved}

We deem a learning curve well-behaved if it shows improved performance with increased training sample sizes, i.e., $\bar{R}_n(A) \ge \bar{R}_{n+1}(A)$ for all $n$.  In slightly different settings, learners that satisfy this property are called smart \cite[page 106]{Devroye1996} and monotone \cite{Viering2019}.  %

\added{There is both experimental and theoretical evidence for well-behaved curves. In the case of large deep learning models, empirical evidence often seems to point to power-law behavior specifically. For problems with binary features and decision trees, exponential curves cannot be ruled out, however.  There generally is no definitive empirical evidence that power laws models are essentially better.  Theory does often point to power-law behavior, but showcases exponentials as well. \added{The most promising theoretical works characterizing the shape are \cite{bousquet2020theory} and \cite{hutter2021learning}.}
Theoretical PA curves are provably monotone, given the problem is well-specified and a Bayesian approaches is used. They favor exponential and power-law shapes as well.}

\subsection{In-Depth Empirical Studies of Parametric Fits}\label{sect:param}

\begin{table}[t]
\caption{Parametric learning curve models.  Note that some curves model performance increase rather than loss decease.  The first column gives the abbreviation used and the number of parameters.  Bold and asterisk marks the paper this model came out as the best fit.}
	\label{table:curve_shapes}
	\centering%
	\begin{tabular}{@{}lllll@{}}
		\toprule
		Reference           & Formula                       & Used in                                                                                                   \\ \midrule
		POW2         & $an^{-b}$                                 & \textbf{\cite{Frey1999}*}\cite{Gu2001}\cite{Singh2005}\cite{Last2007} \\
		POW3       & $an^{-b}+c$                                 & \textbf{\cite{Gu2001}*\cite{Kolachina2012}*}\cite{Cortes1994}\cite{Brumen2012} \\
		LOG2       & $-a \log(n) +c$                      & \textbf{\cite{Singh2005}*}\cite{Frey1999}\cite{Gu2001}\cite{Last2007}\cite{Brumen2012} \\
		EXP3       & $a \exp(-bn) + c$                  & \textbf{\cite{Brumen2012}*}\cite{Kolachina2012}   \\
		EXP2      & $a \exp(-bn)$                            & \cite{Frey1999}\cite{Singh2005}\cite{Last2007}      \\
		LIN2             & $-an+b$                              & \cite{Frey1999}\cite{Singh2005}\cite{Last2007}\cite{Brumen2012}      \\
		VAP3      & $\exp(a + b/n + c \log(n))$                 & \cite{Gu2001}   \\
		MMF4          & $(ab + cn^d)/(b + n^d)$                     & \cite{Gu2001}     \\
		WBL4           & $c - b \exp(-a n^d)$                      & \cite{Gu2001}   \\
		EXP4         & $c - \exp(-an^\alpha +   b)$               & \cite{Kolachina2012}    \\
		EXPP3         & $c - \exp((n-b)^\alpha)$                      & \cite{Kolachina2012}   \\
		POW4          & $c-(-an+b)^{-\alpha}$                    & \cite{Kolachina2012}     \\
		ILOG2         & $c -(a/\log(n))$                         & \cite{Kolachina2012}   \\
		EXPD3      & $c-(c-a)\exp(-bn)$                & \cite{Boonyanunta2004}   \\ \bottomrule                    
	\end{tabular}
\end{table}

Various works have studied the fitting of empirical learning curves and found that they typically can be modelled with function classes depending on few parameters. Table \ref{table:curve_shapes} provides a comprehensive overview of the parametric models studied in machine learning, models for human learning in \cite{anzanello2011learning} may offer further candidates. Two of the primary objectives in studies of fitting learning curves are how \added{well} a model interpolates an empirical learning curve over an observed range of training set sizes and how well it can extrapolate beyond that range.  

Studies investigating these parametric forms often find the power law with offset (POW3 in the table) to offer a good fit. The offset makes sure that a non-zero asymptotic error can be properly modeled, which seems a necessity in any challenging real-world setting.
Surprisingly, even though Frey  and  Fisher \cite{Frey1999} do not include this offset $c$ and use POW2, they find for decision trees that on 12 out of 14 datasets they consider, the power law fits best. Gu  et  al.\ \cite{Gu2001} extend this work to datasets of larger sizes and, next to decision trees, also uses logistic regression as a learner. They use an offset in their power law and consider other functional forms, notably, VAP, MMF, and WBL. For extrapolation, the power law with bias performed the best overall. Also Last \cite{Last2007} trains decision trees and finds the power law to perform best. Kolachina  et  al.\ \cite{Kolachina2012} give learning curves for machine translation in terms of BLUE score for 30 different settings. Their study considers several parametric forms (see Table \ref{table:curve_shapes}) but also they find that the power law is to be preferred.  

Boonyanunta and Zeephongsekul \cite{Boonyanunta2004} perform no quantitative comparison and instead postulate that a differential equation models learning curves, leading them to an exponential form, indicated by EXPD in the table. \cite{Ahmad1988} empirically finds exponential behavior of the learning curve for a perceptron trained with backpropagation on a toy problem with binary inputs, but neither performs an in depth comparison.  In addition, the experimental setup is not described precisely enough: for example, it is not clear how step sizes are tuned or if early stopping is used. 

Three studies find more compelling evidence for deviations from the power law. The first,
\cite{cohn1991can}, can be seen as an in-depth extension of \cite{Ahmad1988}. They train neural networks on four synthetic datasets and compare the learning curves using the $r^2$ goodness of fit. %
Two synthetic problems are linearly separable, the others require a hidden layer, and all can be modeled perfectly by the network. Whether a problem was linearly separable or not doesn't matter for the shape of the curve. For the two problems involving binary features exponential learning curves were found, whereas problems with real-valued features a power law gave the best fit.  However, they also note, 
that it is not always clear that one fit is significantly better than the other. 

The second study, \cite{Singh2005}, evaluates a diverse set of learners on four datasets and shows the logarithm (LOG2) provides the best fit. The author has some reservations about the results and mentions that the fit focuses on the first part of the curve as a reason that the power law may seem to perform worse, besides that POW3 was also not included. Given that many performance measures are bounded, parametric models that increase or decrease beyond any limit should eventually give an arbitrarily bad fit for increasing $n$. As such, LOG2 is anyway suspect. 

The third study, \cite{Brumen2012}, considers only the performance of the fit on training data, e.g. already observed learning curves points. They \emph{do} only use learning curve models with a maximum of three parameters. They employs a total of 121 datasets and use C4.5 for learning. In 86 cases, the learning curve shape fits well with one of the functional forms, in 64 cases EXP3 gave the lowest overall MSE, in 13 it was POW3. %
A signed rank test shows that the exponential outperforms all others models mentioned for \cite{Brumen2012} in Table \ref{table:curve_shapes}. Concluding, the first and last work provide strong evidence against the power law in certain settings.

\added{Finally, in what is probably the most extensive learning curve study to date \cite{lcdb2022}, the authors find power laws to often provide good fits though not necessarily preferable to some of the other models.  In particular, for larger data sets, all four 4-parameter models from Table \ref{table:curve_shapes} often perform on par, but MMF4 and WBL4 outperform EXP4 and POW4 if enough curve data is used for fitting.}

\subsection{Power Laws and Eye-balling Deep Net Results}\label{sect:ohno}

Studies of learning curves of deep neural networks mostly claim to find power-law behavior. However, initial works offer no quantitative comparisons to other parametric forms and only consider plots of the results, calling into question the reliability of such claims. Later contributions find power-law behavior over many orders of magnitude of data, offering \added{somewhat} stronger empirical evidence. 

Sun et al. \cite{Sun2017} state that their \added{mean average precision} performance on a large-scale internal Google image dataset increases logarithmically in dataset size.  This claim is called into question by \cite{Hestness2017} and we would agree: there seems to be little reason to believe this increase follows a logarithm. %
Like for \cite{Singh2005}, that also finds that the logarithm fit well, one should remark that the performance in terms of \added{mean average precision} is always bounded from above and therefore the log model should eventually break.  As opposed to \cite{Sun2017}, \cite{Joulin2016} does observe diminishing returns in a similar large-scale setting. \cite{Mahajan2018} also studies large-scale image classification and find learning curves that level off more clearly in terms of accuracy over orders of magnitudes. They presume that this is due to the maximum accuracy being reached, but note that this cannot explain the observations on all datasets.   In the absence of any quantitative analysis, these results are possibly not more than suggestive of power-law behavior.

\added{Hestness et al.\ \cite{Hestness2017} observe} power laws over multiple orders of magnitude of training set sizes for a broad range of domains: machine translation (error rate), language modeling (cross entropy), image recognition (top-1 and top-5 error rate, cross entropy) and speech recognition (error rate). Its exponent was found to be between $-0.07$ and $-0.35$ and mostly depends on the domain. Architecture and optimizer primarily determine the multiplicative constant. %
For small sample sizes, however, the power law supposedly does not hold anymore, as the neural network converges to a random guessing solutions. %
\added{In addition, one should note that that the work does not consider any competing models nor does it provide an analysis of the goodness of fit.} 
\added{Moreover, to} uncover the power law, significant tuning of the hyperparameters and model size per sample size is necessary, otherwise deviations occur.  \added{Interestingly, \cite{hoiem2021learning} investigates robust curve fitting for the error rate using the power law with offset and find exponents of size $-0.3$ and $-0.7$, thus of larger magnitude than \cite{Hestness2017}.} 

Kaplan  et  al.\ \cite{Kaplan2020} and Rosenfeld  et  al.\ \cite{Rosenfeld2020} \added{also rely on} power laws for image classification and natural language processing \added{and remarks similar to those we made about \cite{Hestness2017} apply}. \cite{Kaplan2020} finds that if the model size and computation time are increased together with the sample size, that the learning curve has this particular behavior. If either is too small this pattern disappears. They find that the test loss also behaves as a power law as function of model size and training time and that the training loss can also be modeled in this way. \cite{Rosenfeld2020} reports that the test loss behaves as a power law in sample size when model size is fixed and vice versa. Both provide models of the generalization error that can be used to extrapolate performances to unseen sample and model sizes and that may reduce the amount of tuning required to get to optimal learning curves.  \added{Some recent additions to this quickly expanding field are \cite{gordon2021data} and \cite{zhai2022scaling}.}

\added{On the whole, given the level of validation provided in these works (e.g. no statistical tests or alternative parametric models, etc.), power laws can currently not be considered more than a good candidate learning curve model.}

\subsection{What Determines the Parameters of the Fit?}\label{sect:what}

Next to the parametric form as such, researchers have investigated what determines the parameters of the fits. \cite{Mukherjee2003}, \cite{Cortes1994}, and \cite{hoiem2021learning} provide evidence that the asymptotic value of the power law and its exponent could be related. Singh \cite{Singh2005} investigates the relation between dataset and classifier but does not find any effect. They do find that the neural networks and \added{the support vector machine (SVM)} are more often well-described by a power law and that decision trees are best predicted by a logarithmic model. Only a limited number of datasets and models was tested however. Perlich et al.\ \cite{Perlich2003} find that the Bayes error is indicative of whether the curves of decision trees and logistic regression will cross or not. In case the Bayes error is small, decision trees will often be superior for large sample sizes. %
All in all, there are few results of this type and most are quite preliminary.

\subsection{Shape Depends on Hypothesis Class} \label{sec_newPAC}

Turning to theoretical evidence, results from learning theory, especially in the form of Probably Approximately Correct (PAC) bounds \cite{vapnik1982estimation,shalev2014understanding}, have been referred to to justify power-law shapes in both the separable and non-separable case \cite{hoiem2021learning}. The PAC model is, however, pessimistic since it considers the performance on a worst-case distribution $P$, while the behavior on the actual fixed $P$ can be much more favorable  (see, for instance, \cite{buntine1989critique,sarrett1989average,haussler1993probably,Haussler1996longer,hestness2017deep}). Even more problematic, the worst-case distribution considered by PAC is \emph{not} fixed, and can depend on the training size $n$, while for the learning curve $P$ is fixed. \added{Thus the actual curve can decrease much faster than the bound (e.g. exponential) \cite{bousquet2020theory}. The curve only has to be beneath the bound but can still also show strange behavior (wiggling).} 

A more fitting approach, pioneered by Schuurmans \cite{schuurmans1995characterizing,schuurmans1997characterizing} and inspired by the findings in \cite{cohn1991can}, does characterize the shape of the learning curve for a fixed unknown $P$. This has also been investigated in \cite{gu2000exponential,gu2000bad} for specific learners. Recently, Bousquet et al. \cite{bousquet2020theory} gave a full characterization of all learning curve shapes for the realizable case and optimal learners. This last work shows that optimal learners can have only three shapes: an exponential shape, a power-law shape, or a learning curve that converges arbitrarily slow. The optimal shape is determined by novel properties of the hypothesis class (not the VC dimension). The result of arbitrarily slow learning is, partially, a refinement of the no free lunch theorem of \cite[Section 7.2]{Devroye1996} and concerns, for example, hypothesis classes that encompass all measurable functions. These results are a strengthening of a much earlier observation by Cover \cite{cover1968rates}.

\subsection{\added{Shape Depends on the Problem}\label{sec:amari}}

\added{There are a number of papers that find evidence for the power-law shape under various assumptions. We also discuss work that finds exponential curves.}

\added{A first provably exponential learning curve can be traced back to the famous work on \added{one-nearest neighbor (1NN)} classifier \cite{cover1967nearest} by Cover and Hart.  They point out that, in a two-class problem, if the classes are far enough apart, 1NN only misclassifies samples from one class if all $n$ training samples are from the other.  Given equal priors, one can show  $\bar{R}_n(A_\text{1NN}) = 2^{-n}$. This suggests that if a problem is well-separated, classifiers can converge exponentially fast. Peterson \cite{peterson1970some} studied 1NN for a two-class problem where $P_X$ is uniform on $[0,1]$ and $P_{Y|X}$ equals $x$ for one of the two classes. In that case of class overlap the curve equals $\tfrac{1}{3}+\tfrac{3 n+5}{2(n+1)(n+2)(n+3)}$, thus we have a slower power law.}

Amari \cite{Amari1993a,Amari1992b} studies (PA) learning curves for a basic algorithm, the Gibbs learning algorithm, in terms of the cross entropy. %
The latter is equal to the logistic loss in the two-class setting and underlies logistic regression.  \cite{Amari1993a} refers to it as the entropic error or entropy loss.  The Gibbs algorithm $A_\text{G}$ is a stochastic learner that assumes a prior over all models considered and, at test time, samples from the posterior defined through the training data, to come to a prediction \cite{opper1991calculation,engel2001statistical}.  Separable data is assumed and the model samples are therefore taken from version space  \cite{engel2001statistical}. 

For $A_G$, the expected cross entropy, $\bar{R}^\text{CE}$, can be shown to decompose using a property of the conditional probability \cite{Amari1993a}. Let $p(S_n)$ be the probability of selecting a classifier from the prior that classifies all samples in $S_n$ correctly. Assume, in addition, that $S_{n} \subset S_{n+1}$. %
Then, while for general losses we end up with an expectation that is generally hard to analyze \cite{Amari1992}, the expected cross entropy simplifies into the difference of two expectations \cite{Amari1993a}: 
\begin{equation}
\bar{R}^\text{CE}_n(A_{G}) = E_{S_{n} \sim P} ~ \log p(S_{n}) - E_{S_{n+1} \sim P} ~ \log p(S_{n+1}).
\end{equation}
Under some additional assumptions, which ensure that the prior is not singular, the behavior asymptotic in $n$ can be fully characterized. Amari \cite{Amari1993a} then demonstrates that
\begin{equation}\bar{R}^\text{CE}_n(A_G) \approx \frac{d}{n} + o\left(\frac{1}{n}\right),\end{equation}
where $d$ is the number of parameters. 

Amari and Murata \cite{Amari1993b} extend this work and consider labels generated by a noisy process, allowing for class overlap. Besides Gibbs, they study algorithms based on maximum likelihood estimation and the Bayes posterior distribution. They find for Bayes and maximum likelihood that the entropic generalization error behaves as $H_0 + \frac{d}{2n}$, while the training error behaves as $H_0 - \frac{d}{2n}$, where $H_0$ is the best possible cross entropy loss. For Gibbs, the error behaves as $H_0 + \frac{d}{n}$, and the training error as $H_0$. In case of model mismatch, the maximum likelihood solution can also be analyzed.  In that setting, the number of parameters $d$ changes to a quantity indicating the number of effective parameters and $H_0$ becomes the loss of the model closest to the groundtruth density in terms of KL-divergence. 

In a similar vein, Amari et al.\ \cite{Amari1992} analyze the 0-1 loss, i.e., the error rate, under a so-called annealed approximation \cite{watkin1993statistical,engel2001statistical}, which approximates the aforementioned hard-to-analyze risk. Four settings are considered, two of which are similar to those in \cite{Amari1993b,Amari1992b,Amari1993a}. The variation in them stems from differences in assumptions about how the labeling is realized, ranging from a unique, completely deterministic labeling function to multiple, stochastic labelings.  Possibly the most interesting result is for the realizable case where multiple parameter settings give the correct outcome or, more precisely, where this set has nonzero measure.  In that case, the asymptotic behavior is described as a power law with an exponent of $-2$. Note that this is essentially faster than what the typical PAC bounds can provide, which are exponents of $-1$ and $-\tfrac{1}{2}$ (a discussion of those results is postponed to section \ref{sec_PAC}).  This possibility of a more rich analysis is sometimes mentioned as one of the reasons for studying learning curves \cite{Seung1992,Haussler1996longer,engel2001statistical}. 

For some settings, exact results can be obtained. If one considers a 2D input space where the marginal $P_X$ is a Gaussian distribution with mean zero and identity covariance, and one assumes a uniform prior over the true linear labeling function, the PA curve for the zero one loss can exactly be computed to be of the form $\frac{2}{3n}$, while, the annealed approximation gives $\frac{1}{n}$ \cite{Amari1992}.  

Schwartz  et  al.\ \cite{Schwartz1990a} use tools similar to Amari's to study the realizable case where all variables (features and labels) are binary and Bayes rule is used. Under their approximations, the PA curve can be completely determined from a histogram of generalization errors of models sampled from the prior.  For large sample sizes, the theory predicts that the learning curve actually has an exponential shape. The exponent depends on the gap in generalization error between the best and second best model. In the limit of a gap of zero size, the shape reverts to a power law. The same technique is used to study learning curves that can plateau before dropping off a second time \cite{tishby1989consistent}. \cite{Levin1989} proposes extensions dealing with label noise. \cite{Seung1992} casts some doubt on the accuracy of these approximations and the predictions of the theory of Schwartz et al. indeed deviate quite a bit from their simulations on toy data. %

\added{Recently, \cite{hutter2021learning} presented a simple classification setup on a countably infinite input space for which the Bayes error equals zero.  The work shows that the learning curve behavior crucially depends on the distribution assumed over the discrete feature space and most often shows power-law behavior.  More importantly, it demonstrates how exponents other than the expected $-1$ or $-\tfrac{1}{2}$ can emerge, which could explain the diverse exponents mentioned in Section \ref{sect:ohno}.} 

\subsection{Monotone Shape if Well-Specified (PA)} \label{sub:pa_monotone}

The PA learning curve is monotone if the prior and likelihood model are correct and Bayesian inference is employed.  This is a consequence of the total evidence theorem \cite{savage1954foundations,Good1967,grunwald2004ignorance}. It states, informally, that one obtains the maximum expected utility by taking into account all observations. %
However, a monotone PA curve does not rule out that the learning curve for individual problems can go up, even if the problem is well-specified, as the work covered in Section \ref{sec:perfect-prior} points out. Thus, if we only evaluate in terms of the learning curve of a single problem, using all data is not always the rational strategy.

It may be of interest to note here that, next to Bayes' rule, there are other ways of consistently updating one's belief---in particular, so-called probability kinematics---that allow for an alternative decision theoretic setting in which a total evidence theorem also applies \cite{graves1989total}.

Of course, in reality, our model probably has some misspecification, which is a situation that has been considered for Gaussian process models and Bayesian linear regression. Some of the unexpected behavior this can lead to is covered in Subsection \ref{sub_misspecified_GP} for PA curves and Subsection \ref{sec:safebayes} for the regular learning curve. %
Next, however, we cover further results in the well-behaved setting.  We do this, in fact, specifically for Gaussian processes. %

\section{Gaussian Process Learning Curves} \label{sec:GP}

In Gaussian process (GP) regression \cite{Rasmussen2006}, it is especially the PA learning curve (Subsection \ref{sub:def}) for the squared loss, under the assumption of a Gaussian likelihood, that has been studied extensively. A reason for this is that many calculations simplify in this setting.  We cover various approximations, bounds, and ways to compute the PA curve.  We also discuss assumptions and their corresponding learning curve shapes and cover the factors that affect the shape. It appears difficult to say something universally about the shape, besides that it cannot decrease faster than $O(n^{-1})$  asymptotically. The fact that these PA learning curves are monotone in the correctly specified setting can, however, be exploited to derive generalization bounds. We briefly cover those as well.  This section is limited to correctly specified, well-behaved curves, Subsection \ref{sub_misspecified_GP} is devoted to ill-behaved learning curves for misspecified GPs.

The main quantity that is visualized in the PA learning curve of GP regression is the Bayes risk or, equivalently, the problem averaged squared error. In the well-specified case, this  is equal to the posterior variance $\sigma_*^2$  \cite[Equation~(2.26)]{Rasmussen2006}\cite{Williams2000},
\begin{equation}
\sigma_*^2 = k(x_*, x_*) - k_*^T (K(X_n, X_n) + \sigma^2 I)^{-1} k_*. \label{eq_pv_reg}
\end{equation}
Here $k$ is the covariance function or kernel of the GP, $K(X_n,X_n)$ is the $n \times n$ kernel matrix of the input training set $X_n$, where $K^{lm}(X_n,X_n) = k(x_l, x_m)$. Similarly, $k_*$ is the vector where the $l$th component is $k(x_l, x_*)$. Finally, $\sigma$ is the noise level assumed in the Gaussian likelihood.

The foregoing basically states that the averaging over all possible problems $\mathcal{P}$ (as defined by the GP prior) and all possible output training samples is already taken care of by Equation \eqref{eq_pv_reg}.  Exploiting this equality, what is then left to do to get to the PA learning curve is an averaging over all test points $x_*$  according to their marginal $P_X$ and an averaging over all possible input training samples $X_n$.  

Finally, for this section, it turns out to be convenient to introduce a notion of PA learning curves in which only the averaging over different input training sets $X_n \in \mathcal{X}^n$ has \emph{not} been carried out yet.  We denote this by $R^\text{PA}(X_n)$ and we will not mention the learning algorithm $A$ (GP regression). %

\subsection{Fixed Training Set and Asymptotic Value} \label{sec:fixed_trn_set}

The asymptotic value of the learning curve and the value of the PA curve for a fixed training set can be expressed in terms of the eigendecomposition of the covariance function.  This decomposition is often used when approximating or bounding GPs' learning curves \cite{Sollich2002,Williams2000,Lederer2019}. %

With $P_X$ be the marginal density and $k(x,x')$ the covariance function. The eigendecomposition constitutes all eigenfunctions $\phi_i$ and eigenvalues $\lambda_i$ that solve for
\begin{equation}
\int k(x,x') \phi(x) dP_X(x) = \lambda \phi(x').
\end{equation}
Usually, the eigenfunctions are chosen so that
\begin{equation}
\int \phi_i(x) \phi_j(x) dP_X(x) = \delta_{ij}, \label{eq:phi_orthonormaal}
\end{equation}
where $\delta_{ij}$ is Kronecker's delta \cite{Rasmussen2006}.  The eigenvalues are non-negative and assumed sorted from large to small ($\lambda_1 \geq \lambda_2 \geq \ldots$). Depending on $P_X$ and the covariance function $k$, the spectrum of eigenvalues may be either degenerate, meaning there may be finite nonzero eigenvalues, or nondegenerate in which case there are infinite nonzero eigenvalues. For some, analytical solutions to the eigenvalues and functions are known, e.g. for the squared exponential covariance function and Gaussian $P_X$. For other cases the eigenfunctions and eigenvalues can be approximated \cite{Rasmussen2006}. 

Now, take $\Lambda$ the diagonal matrix with $\lambda_i$ on the diagonal and let $\Phi_{li} = \phi_i(x_l)$, where each $x_l$ comes from the training matrix $X_n$. The dependence on the training set is indicated by $\Phi = \Phi(X_n)$. %
For a fixed training set, the squared loss over all problems can then be written as 
\begin{equation}
R^\text{PA}(X_n) = \Tr(\Lambda^{-1} + \sigma^{-2} \Phi(X_n)^T \Phi(X_n))^{-1}, \label{GP_exact_eigen}
\end{equation}
which is exact \cite{Sollich2002}. The only remaining average to compute to come to a PA learning curve is with respect to $X_n$. This last average is typically impossible to calculate analytically (see \cite[p. 168]{Rasmussen2006} for an exception). This leads one to consider the approximations and bounds covered in Section \ref{sect:approx}. 

The asymptotic value of the PA learning curve is \cite{Gratiet2015}
\begin{equation}\label{eq:asymp}
\bar{R}^\text{PA}_\infty = \sum_{i = 1}^\infty \frac{\lambda_i \tau}{\tau + \lambda_i},
\end{equation}
where convergence is in probability and almost sure for degenerate kernels. Moreover, it is assumed that $\sigma^2 = n \tau$ for a constant $\tau$, which means that the noise level grows with the sample size.  The assumption seems largely a technical one and Le Gratiet and Garnier \cite{Gratiet2015} claim this assumption can still be reasonable in specific settings.

\subsection{Two Regimes, Effect of Length Scale on Shape}

For many covariance functions (or kernels), there is a characteristic length scale $l$ that determines the distance in feature space one after which the regression function can change significantly.   Williams  and  Vivarelli \cite{Williams2000} make the qualitative observation that this lead the PA curve to often have two regimes: initial and asymptotic.  If the length scale of the GP is not too large, the initial decrease of the learning curve is approximately linear in $n$ (initial regime). They explain that, initially, the training points are far apart, and thus they can almost be considered as being observed in isolation. Therefore, each training point reduces the posterior variance by the same amount, and thus the decrease is initially linear. However, when $n$ is larger, training points get closer together and their reductions in posterior variance interact. Then reduction is not linear anymore because an additional point will add less information than the previous points. Thus there is an effect of diminishing returns and the decrease gets slower: the asymptotic regime \cite{Sollich2002,Williams2000}.

The smaller the length scale, the longer the initial linear trend, because points are comparatively further apart  \cite{Williams2000,Sollich2002}. Williams and Vivarelli further find that changing the length scale effectively rescales the amount of training data in the learning curve for a uniform marginal $P_X$. Furthermore, they find that a higher noise level and smoother covariance functions results in earlier reaching of the assymptotic regime for the uniform distribution. It remains unclear how these results generalize to other marginal distributions. 

Sollich and Halees \cite{Sollich2002} note that in the asymptotic regime, the noise level has a large influence on the shape of the curve since here the error is reduced by averaging out noise, while in the non-asymptotic regime the noise level hardly plays a role. They always assume that the noise level is much smaller than the prior variance (the expected fluctuations of the function before observing any samples). Under that assumption, they compare the error of the GP with the noise level to determine the regime: if the error is smaller than the noise, it indicates that one is reaching the asymptotic regime. 

\subsection{Approximations and Bounds}\label{sect:approx}

A simple approximation to evaluate the expectation with respect to the training set from Equation \ref{GP_exact_eigen} is to replace the matrix $\Phi(X_n)^T \Phi(X_n)$ by its expected value $nI$ (this is the expected value due to Equation \ref{eq:phi_orthonormaal}). %
This results in 
\begin{equation}
\bar{R}^\text{PA}_n \approx  \sum_{i = 1}^\infty \frac{\lambda_i\sigma^2}{\sigma^2 + n \lambda_i}. \label{GP_naive}
\end{equation}
The approximation, which should be compared to Equation \eqref{eq:asymp}, is shown to be an upper bound on the training loss and a lower bound on the PA curve\cite{Opper1998}. From asymptotic arguments, it can be concluded that the PA curve cannot decrease faster than $\tfrac{1}{n}$ for large $n$ \cite[p. 160]{Rasmussen2006}.
Since asymptotically the training and test error coincide, Opper  and  Vivarelli \cite{Opper1998} expect  this approximation to give the correct asymptotic value, which, indeed, is the case \cite{Gratiet2015}. %

The previous approximation works well for the asymptotic regime but for non-asymptotic cases it is not accurate. Sollich \cite{Sollich1998} aims to approximate the learning curve in such a way that it can characterizes both regimes well.  To that end, he, also in collaboration with Halees \cite{Sollich2002}, introduces three approximations based on a study of how the matrix inverse in Equation \eqref{GP_exact_eigen} changes when $n$ is increased. He derives recurrent relations that can be solved and lead to upper and lower approximations that typically enclose the learning curve. The approximations have a form similar to those in Equation \eqref{GP_naive}. While \cite{Sollich1998} initially  hypothesized these approximations could be actual bounds on the learning curve, Sollich and Halees \cite{Sollich2002} gave several counter examples and disproved this claim. %

For the noiseless case, in which $\sigma=0$, Michelli and Wahba \cite{Michelli1981} give a lower bound for $R^\text{PA}(X_n)$, which in turn provides a lower bound on the learning curve: 
\begin{equation}
\bar{R}^\text{PA}_n \geq \sum_{i=n+1}^\infty \lambda_i. \label{eq_weggehaalde_paragraaf}
\end{equation}
Plaskota \cite{Plaskota1996} extends this result to take into account noise in the GP and %
Sollich and Halees \cite{Sollich2002} provide further extensions that hold under less stringent assumptions. 
Using a finite-dimensional basis, \cite{Trecate1999} develops a method to approximate GPs that scales better in $n$ and finds an upper bound. The latter immediately implies a bound on the learning curve as well.

Several alternative approximations exist. For example, S{\"a}rkk{\"a} \cite{Sarkka2011} uses numerical integration to approximate the learning curve, for which the eigenvalues do not need to be known at all. %
Opper \cite{Opper1997} gives an upper bound for the entropic loss using techniques from statistical physics and derives the asymptotic shape of this bound for Wiener processes for which $k(x,x') = \min(x,x')$ and the squared exponential covariance function. %
\cite{Sollich2002} notes that in case the entropic loss is small it approximates the squared error of the GP, thus for large $n$ this also implies a bound on the learning curve.

\subsection{Limits of the Eigenvalue Spectrum}

Sollich  and  Halees \cite{Sollich2002} also explore what the limits are of bounds and approximations based on the eigenvalue spectrum. They create problems that have equal spectra but different learning curves, indicating that learning curves cannot be predicted reliably based on eigenvalues alone. 
Some works rely on more information than just the spectrum, such as  \cite{Williams2000,Trecate1999} whose bounds also depend on integrals involving the weighted and squared versions of the covariance function. Sollich and Halees also shows several examples where the lower bounds from \cite{Opper1998} and \cite{Plaskota1996} can be arbitrarily loose, but at the same time cannot be significantly tightened. Also their own approximation, which empirically is found to be the best, cannot be further refined. These impossibility results leads them to question the use of the eigenvalue spectrum to approximate the learning curve and what further information may be valuable. 

Malzahn  and  Opper \cite{Malzahn2001} re-derive the approximation from \cite{Sollich2002} using a different approach. Their variational framework may provide more accurate approximations to the learning curve, presumably also for GP classification \cite{Sollich2002,Rasmussen2006}. %
It has also been employed, among others, to estimate the variance of PA learning curves \cite{Malzahn2003,Malzahn2001c}. %

\subsection{Smoothness and Asymptotic Decay} \label{sec:smooth}

The smoothness of a GP can, in 1D, be characterized by the Sacks-Ylvisaker conditions \cite{ritter1995multivariate}.  These conditions capture an aspect of the smoothness of a process in terms of its derivatives.  The order $s$ used in these regularity conditions indicates, roughly, that the $s$th derivative of the stochastic process exists, while the $(s+1)$th does not.  Under these conditions, Ritter \cite{Ritter1996}  showed that the asymptotic decay rate of the PA learning curve is of order $O(n^{-(2s+1)/(2s+2)})$. Thus the smoother the process the faster the decay rate. %

As an illustration: the squared exponential covariance induces a process that is very smooth as all orders of derivatives exist. The smoothness of the so-called modified Bessel covariance function \cite{Rasmussen2006} is determined by its order $k$, which relates to the order in the Sacks-Ylvisaker conditions as $s=k-1$. If $k=1$, the covariance function leads to a process that is not even once differentiable. In the limit of $k \rightarrow \infty$, it converges to the squared exponential covariance function \cite{Williams2000,Rasmussen2006}. For the roughest case $k=1$ the learning curve behaves asymptotically as $O(n^{-1/2})$. %
For the smoother SE covariance function the rate is $O(n^{-1} \log(n))$ \cite{Opper1997}. %
For other rates see \cite[Chapter~7]{Rasmussen2006} and \cite{Williams2000}.

In particular cases, the approximations in \cite{Sollich2002} can have stronger implications for the rate of decay. If the eigenvalues decrease as a power law $\lambda_i \sim i^{-r}$, in the asymptotic regime ($\bar{R}^\text{PA}_n \ll \sigma^2$), the upper and lower approximations coincide and predict that the learning curve shape is given by $({n}/{\sigma^2})^{-(r-1)/r}$. For example, for covariance function of the classical Ornstein–Uhlenbeck process, i.e., $k(x,x') = \exp - \tfrac{|x-x'|}{\ell}$ with $\ell$ the length scale, we have $\lambda_i \sim i^{-2}$. The eigenvalues of the squared exponential covariance function decay faster than any power law.  Taking $r \rightarrow \infty$, this implies a shape of the form $\frac{\sigma^2}{n}$. These approximation are indeed in agreement with known exact results \cite{Sollich2002}. In the initial regime ($\bar{R}^\text{PA}_n \gg \sigma^2$), if one takes $\sigma \rightarrow 0$ the lower approximation gives $n^{-(r-1)}$. In this case, the suggested curve for the Ornstein–Uhlenbeck process takes on the form $n^{-1}$, which agrees with exact calculations as well. In the initial regime for the squared exponential, no direct shape can be computed without additional assumptions. Assuming $d=1$, a uniform input distribution, and large $n$, the approximation suggests a shape of the form $n e^{-cn^2}$ for some constant $c$ \cite{Sollich2002}, which is faster even than exponential.

\subsection{Bounds through Monotonicity}

Subsection \ref{sub:pa_monotone} indicated that the PA learning curve is always monotone if the problem is well specified.  Therefore, under the assumption of a well-specified GP, its learning curve is monotone when its average is considered over all possible training sets (as defined by the correctly specified prior). As it turns out, however, this curve is already monotone \emph{before} averaging over all possible training sets, as long as the smaller set is contained in the larger, i.e., $R^\text{PA}(X_n) \ge R^\text{PA}(X_{n+1})$ for all $X_n \subset X_{n+1}$.   This is because the GP's posterior variance decreases with every addition of a training object. This can be proven from standard results on the conditioning a multivariate Gaussians \cite{Williams2000}. Such sample-based monotonicity of the posterior variance can generally be obtained if the likelihood function is modeled by an exponential family and a corresponding conjugate prior is used \cite{Al-Saleh2003}.  

Williams  and  Vivarelli  \cite{Williams2000} use this result to construct bounds on the learning curve.  The key idea is to take the $n$ training points and treat them as $n$ training sets of just a single point. Each one gives an estimate of generalization error for the test points (Equation \ref{eq_pv_reg}). Because the error always decreases when increasing the training set, the minimum over all train points creates a bound on the error for the whole training set for the test point. The minimizer is the closest training point. %

To compute the a bound on the generalization error, one needs to consider the expectation w.r.t. $P_X$. In order to keep the analysis tractable, Williams  and  Vivarelli  \cite{Williams2000} limit themselves to 1D input spaces where $P_X$ is uniform and perform the integration numerically. They further refine the technique to training sets of two points as well, and, find that as expected the bounds become tighter, since larger training sets imply better generalization. Sollich  and  Halees \cite{Sollich2002} extend their technique to non-uniform 1D $P_X$.  By considering training points in a ball of radius $\rho$ around a test point, Lederer et al.\ \cite{Lederer2019}  derive a similar bound that converges to the correct asymptotic value. In contrast, the original bound from \cite{Williams2000} becomes looser as $n$ grows. %
Experimentally, \cite{Lederer2019} shows that under a wide variety of conditions their bound is relatively tight.

\section{Ill-Behaved Learning Curves}\label{sect:ill}

It is important to understand that learning curves do not always behave well and that this is not necessarily an artifact of the finite sample or the way an experiment is set up.  Deterioration with more training data can obviously occur when considering the curve $R(A(S_n))$ for a particular training set, because for every $n$, we can be unlucky with our draw of $S_n$. That ill-behavior can also occur in expectation, i.e., for $\bar{R}_n(A)$, may be less obvious.

In the authors' experience, most researchers expect improved performance of their learner with more data.  Less anecdotal evidence can be found in literature. \cite[page~153]{shalev2014understanding} states that when $n$ surpasses the VC-dimension, the curve must start decreasing. \cite[Subsection~9.6.7]{Duda2012} claims that for many real-world problems they decay monotonically.  \cite{Tax2008} calls it expected that performance improves with more data and \cite{Gu2001} makes a similar claim, \cite{Meng2014} and \cite{Ting2017} consider it conventional wisdom, and \cite{Boonyanunta2004} considers it widely accepted. Others assume well-behaved curves \cite{Provost1999}, which means that curves are smooth and monotone \cite{weiss2014generating}. %
\added{Note that a first example of nonmonotonic behavior had actually already been given in 1989 \cite{vallet1989linear}, which we further cover in Section \ref{sub:peaking}.}

\begin{figure}[tbh]
	\centering
	\includegraphics[width=0.42\textwidth]{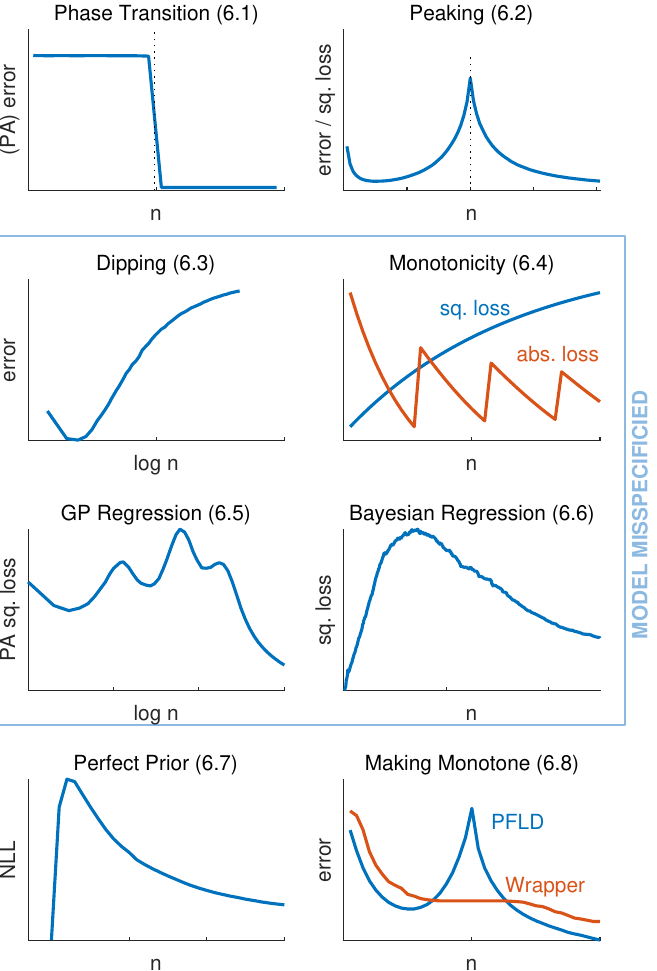}
	\caption{Qualitative overview of various learning curve shapes placed in different categories with references to their corresponding subsections. All have the sample size $n$ on the horizontal axis. %
	Dotted lines indicate the transition from under to overparametrized models. Abbreviations; error: classification error; sq.~loss: squared loss; NLL: negative log likelihood; abs.~loss: absolute loss; PA indicates the problem-average learning curve is shown.}
	\label{fig:overview}
\end{figure}

Before this section addresses actual bad behavior, we cover phase transitions, which are at the brink of becoming ill-behaved.   Possible solutions to nonmonotonic behavior are discussed at the end. Figure~\ref{fig:overview} provides an overview of types of ill-behaved learning curve shapes with subsection references. The code reproducing these curves (all based on actual experiments) can be retrieved from \url{https://github.com/tomviering/ill-behaved-learning-curves}.

\subsection{Phase Transitions}

As for physical systems, in a phase transition, particular learning curve properties change relatively abruptly, (almost) discontinuously. Figure \ref{fig:overview} \added{under (6.1)} gives an example of how this can manifest itself. In learning, techniques from statistical physics can be employed to model and analyze these transitions, where it typically is studied in the limit of large samples and high input dimensionality  \cite{engel2001statistical}.  Most theoretical insights are limited to relatively simple learners, like the perceptron, and often apply to PA curves.

Let us point out that abrupt changes also seem to occur in human learning curves \cite{bryan1897studies,bryan1899studies}, in particular when the task is complex and has a hierarchical structure \cite{vetter1997phase}. A first mention of the occurrence of phase transitions, explicitly in the context of learning curves, can be found in \cite{patarnello1987learning}. It indicates the transition from memorization to generalization, which occurs, roughly, around the time that the full capacity of the learner has been used.  Gy{\"{o}}rgyi \cite{Gyorgyi1990} provides a first, more rigorous demonstration within the framework of statistical physics---notably, the so-called thermodynamic limit \cite{engel2001statistical}.  In this setting,  actual transitions happen for single-layer perceptrons where weights take on binary values only.

The perceptron and its thermodynamic limit are considered in many later studies as well. The general finding is that, when using discrete parameter values---most often binary weights, phase transitions can occur \cite{Watkin1993,Seung1992,Kang1993}. The behavior is often characterized by long plateaus where the perceptron cannot learn at all (usually in the overparametrized, memorization phase, where $n < d$) and has random guessing performance, until a point where the perceptron starts to learn (at $n > d$, the underparametrized regime) at which a jump occurs to non-trivial performance. %

Phase transitions are also found in two-layer networks with binary weights and activations \cite{Opper1995,Kang1993,Schwarze1993}. This happens for the so-called  parity problem where the aim is to detect the parity of a binary string \cite{Hansel1992} for which Opper \cite{Opper2001} found phase transitions in approximations of the learning curve. %
Learning curve bounds may display phase transitions as well \cite{Seung1995,Haussler1996longer},  though  \cite{Seung1992,Seung1995} question whether these will also occur in the actual learning curve. Both Sompolinsky \cite{sompolinsky1993theoretical} and Opper \cite{Opper2001} note that the sharp phase transitions predicted by theory, will be more gradual in real-world settings. Indeed, when studying this literature, one should be careful in interpreting theoretical results, as the transitions may occur only under particular assumptions or in limiting cases. 

For unsupervised learning, phase transitions have been shown to occur as well \cite{biehl1993statistical,hoyle2007statistical} (see the latter for additional references).  Ipsen and Hansen \cite{ipsen2019phase} extend these analyses to PCA with missing data.  They also show phase transitions in experiments on real-world data sets.  \cite{bhat2014adapting} provides one of the few real application papers where a distinct, intermediate plateau is visible in the learning curve.  

For Figure \ref{fig:overview} \added{under (6.1)}, we constructed a simple phase transition based on a two-class classification problem, $y\in\{+1,-1\}$, with the first 99 features standard normal and the 100th feature set to $\frac{y}{100}$.  PFLD's performance shows a transition at $n=100$ for the error rate.

\subsection{Peaking and Double Descent} \label{sub:peaking}

The term peaking indicates that the learning curve takes on a maximum, typically in the form of a cusp, see Figure \ref{fig:overview} \added{under (6.2)}. Unlike many other ill behaviors, peaking can occur in the realizable setting. Its cause seems related to instability of the model.   This peaking should not be confused with peaking for feature curves as covered in Subsection \ref{sect:fc}, which is related to the curse of dimensionality. %
Nevertheless, the same instability that causes peaking in learning curves can also lead to a peak in feature curves, see Figure \ref{fig:surfaceplot}. The latter phenomenon has gained quite some renewed attention under the name double descent \cite{Belkin2019}.

By now, the term (sample-wise) double descent has become a term for the  peak in the learning curve for deep neural networks \cite{nakkiran2019deep,Nakkiran2019more}. Related terminologies are model-wise double descent, that describe a peak in the plot of performance versus model size, and epoch-wise double descent, that shows a peak in the training curve \cite{nakkiran2019deep}.

Peaking was first observed for \added{the pseudo-Fisher's linear discriminant (PFLD)} \cite{vallet1989linear} and has been studied already for quite some time \cite{loog2020brief}. The PFLD is the classifier minimizing the squared loss, using minimum-norm or ridgeless linear regression based on the pseudo-inverse. %
PFLD often peaks at $d \approx n$, both for the squared loss and  classification error.  A first theoretical model explaining this behavior in the thermodynamical limit is given in \cite{opper1990ability}. In such works, originating from statistical physics, the usual quantity of interest is $\alpha = \tfrac{d}{n}$ that controls the relative sizes for $d$ and $n$ going to infinity  \cite{opper1991calculation,watkin1993statistical,engel2001statistical}.

Raudys and Duin \cite{Raudys1998} investigate this behavior in the finite sample setting where each class is a Gaussian. They approximately decompose the generalization error in three terms. The first term measures the quality of the estimated means and the second the effect of reducing the dimensionality due to the pseudo-inverse. These terms reduce the error when $n$ increases. The third term measures the quality of the estimated eigenvalues of the covariance matrix. This term increases the error when $n$ increases, because more eigenvalues need to be estimated at the same time if $n$ grows, reducing the quality of their overall estimation. These eigenvalues are often small and as the model depends on their inverse, small estimation errors can have a large effect, leading to a large instability \cite{skurichina1996stabilizing} and peak in the learning curve around $n \approx d$.  Using an analysis similar to \cite{Raudys1998}, \cite{Krijthe2016} studies the peaking phenomenon in semi-supervised learning  (see \cite{chapelle2010semi}) and shows that unlabeled data can both mitigate or worsen it.

Peaking of the PFLD can be avoided through regularization, e.g. by adding $\lambda I$ to the covariance matrix \cite{Raudys1998,skurichina1996stabilizing}.  The performance of the model is, however, very sensitive to the correct tuning of the ridge parameter $\lambda$ \cite{Tresp2005,skurichina1996stabilizing}. Assuming the data is isotropic, \cite{nakkiran2020optimal} shows that peaking disappears for the optimal setting of the regularization parameter. 
Other, more heuristic solutions change the training procedure altogether, e.g., \cite{Duin1995} uses an iterative procedure that decides which objects PFLD should be trained on, as such reducing $n$ and removing the peak. \cite{Skurichina1999} adds copies of objects with noise, increasing $n$, or increases the dimensionality by adding noise features, increasing $d$. Experiments show this can remove the peak and improve performance. %

Duin \cite{Duin2000} illustrates experimentally that the SVM may not suffer from peaking in the first place. Opper \cite{Opper2001} suggests a similar conclusion based on a simplistic thought experiment.  For specific learning problems, both \cite{opper1990ability} and \cite{Watkin1993} already give a theoretical underpinning for the absence of double descent for the perceptron of optimal (or maximal) stability, which is a classifier closely related to the SVM. Opper \cite{opper2001universal} studies the behavior of the SVM in the thermodynamic limit which does not show peaking either. Spigler et al. \cite{spigler2019jamming} show, however, that double descent for feature curves can occur using the (squared) hinge loss, where the peak is typically located at an $n>d$.

Further insight of when peaking can occur may be gleaned from recent works like \cite{advani2017high} and  \cite{hastie2019surprises}, which perform a rigorous analysis of the case of Fourier Features with PFLD using random matrix theory.  Results should, however, be interpreted with care as these are typically derived in an asymptotic setting where both $n$ and $d$ (or some more appropriate measure of complexity) go to infinity, i.e., a setting similar to the earlier mentioned thermodynamic limit. Furthermore, \cite{d2020triple} shows that a peak can occur where the training set size $n$ equals the input dimensionality $d$, but also when $n$ matches the number of parameters of the learner, depending on the latter's degree of nonlinearity.  Multiple peaks are also possible for $n<d$  \cite{nakkiran2020optimal}.

\subsection{Dipping and Objective Mismatch}\label{sect:dip}

\begin{figure}[tb]
 \centering
	\includegraphics[width=0.84\columnwidth]{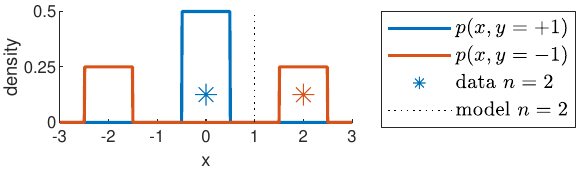}
	\caption{\added{A two-class problem that dips for various linear classifiers  (cf.~\cite{Loog2012}). The sample data (the two $\convolution$s) shows that, with small samples,  the linear model with optimal error rate can be obtained.  However, due to the surrogate loss typically optimized, the decision boundary obtained in the infinite sample limit is around the suboptimal $x=0$.}}
	\label{fig:dipping}
\end{figure}

In dipping, the learning curve may initially improve with more samples, but the performance eventually deteriorates and never recovers, even in the limit of infinite data \cite{Loog2012}, see Figure \ref{fig:overview} \added{below (6.3)}. Thus the best expected performance is reached at a finite training set size.  By constructing an explicit problem \cite[page 106]{Devroye1996}, Devroye et al.\ already showed that the nearest neighbor classifier is not always smart, meaning its learning curve can go up locally.  A similar claim is made for kernel rules \cite[Problems 6.14 and 6.15]{Devroye1996}.  A 1D toy problem for which many well-known linear classifiers (e.g., SVM, logistic regression, LDA, PFLD) dip is given in Figure \ref{fig:dipping}.  In a different context, Ben-David et al.\ \cite{Ben-David2012} provide an even stronger example where all linear classifiers optimizing a convex surrogate loss converge in the limit to the worst possible classifier for which the error rate approaches 1.
Another example, Lemma 15.1 in \cite{Devroye1996}, gives an insightful case of dipping for likelihood estimation.

What is essential for dipping to occur is that the classification problem at hand is misspecified, and that the learner optimizes something else than the evaluation metric of the learning curve.  Such \emph{objective misspecification} is standard since many evaluation measures such as error rate, AUC, F-measure, and so on, are notoriously hard to optimize (cf.~e.g., \cite[page~119]{shalev2014understanding}\cite{loog2016measuring}). %
If classification-calibrated loss functions are used and the hypothesis class is rich enough to contain the true model, then minimizing the surrogate loss will also minimize the error rate \cite{bartlett2004large,Ben-David2012}. %
\added{As such, consistent learners, that deliver asymptotically optimal performance by definition, cannot have learning curves that keep increasing monotonically and, therefore, cannot dip.} %

Other works also show dipping of some sort. For example, Frey and Fisher \cite{Frey1999} fit C4.5 to a synthetic dataset that has binary features for which the parity of all features determines the label. When fitting C4.5 the test error increases with the amount of training samples. They attribute this to the fact that the C4.5 is using a greedy approach to minimize the error, and thus is closely related to objective misspecification. 
Brumen et al.\ \cite{Brumen2012} also shows an ill-behaving curve of C4.5 that seems to go up. They note that 34 more curves could not be fitted well using their parametric models, where possibly something similar is going on. In \cite{vanschoren2008learning}, we find another potential example of dipping as, in Figure 6, the accuracy goes down with increasing sample sizes. 

Anomaly or outlier detection using $k$-nearest neighbors ($k$NN) can also shows dipping behavior \cite{Ting2017} (referred to as gravity defying learning curves). Also here is a mismatch between the objective that is evaluated with, i.e., the AUC, and $k$NN that does not optimize the AUC. %
Hess and Wei \cite{Hess2010} also show $k$NN learning curves that deteriorate in terms of AUC in the standard supervised setting.   %

Also in active learning \cite{settles2009active} for classification, where the test error rate is often plotted against the size of the (actively sampled) training set, learning curves are regularly reported to dip  \cite{schohn2000less,konyushkova2015introducing}.  In that case, active learners provide optimal performance for a number of labeled samples that is smaller than the complete training set.  This could be interpreted as a great success for active learning.  It implies that even in regular supervised learning, one should maybe use an active learner to pick a subset from one's complete training set, as this can improve performance. It cannot be ruled out, therefore, that the active learner uses an objective that matches better with the evaluation measure \cite{loog2016empirical}. 

Meng and Xie \cite{Meng2014} construct a dipping curve in the context of time series modeling with ordinary least squares. In their setting, they use an adequate parametric model, but the distribution of the noise changes every time step, which leads least squares to dipping. In this case, using the likelihood to fit the model resolves the non-monotonicity.

Finally, so-called negative transfer \cite{wang2019characterizing}, as it occurs in transfer learning and domain adaptation \cite{weiss2016survey,kouw2019review}, can be interpreted as dipping as well. In this case, more source data deteriorates performance on the target and the objective mismatch stems from the combined training from source and target data instead of the latter only.  

\subsection{Risk Monotonicity and ERM}\label{sect:mono}

Several novel examples of non-monotonic behavior for density estimation, classification, and regression by means of standard empirical risk minimization (ERM) are shown in \cite{loog2019minimizers}. Similar to dipping, the squared loss increases with $n$, but in contrast does eventually recover, see Figure \ref{fig:overview} \added{under (6.4)}. However, these examples cannot be explained either in terms of dipping or peaking.  Dipping is ruled out as, in ERM, the learner optimizes the loss that is used for evaluation.  In addition, non-monotonicity can be demonstrated for any $n$ and so there is no direct link with the capacity of the learner, ruling out an explanation in terms of peaking.

Proofs of non-monotonicity are given for squared, absolute, and hinge loss.  It is demonstrated that likelihood estimators suffer the same deficiency. Two learners are reported that are provably monotonic: mean estimation based on the $L_2$ loss and the memorize algorithm. The latter algorithm does not really learn but outputs the majority voted classification label of each object if it has been seen before. Memorize is not PAC learnable \cite{shalev2014understanding,engel2001statistical}, illustrating that monotonicity and PAC are essentially different concepts.  \added{Here, we like to mention \cite{power2022grokking}, which shows generalization even beyond memorization of the finite training data.}

It is shown experimentally that regularization can actually worsen the non-monotonic behavior.  In contrast, Nakkiran \cite{nakkiran2020optimal} shows that optimal tuning of the regularization parameter can guarantee monotonicity in certain settings. A final experiment from \cite{loog2019minimizers} shows a surprisingly jagged learning curve for the absolute loss, see Figure \ref{fig:overview} \added{under (6.4)}. 

\subsection{Misspecified Gaussian Processes} \label{sub_misspecified_GP}

Gaussian process misspecification has been studied in the regression setting where the so-called teacher model provides the data, while the student model learns, assuming a covariance or noise model different from the teacher. If they are equal, the PA curve is monotone (Section \ref{sub:pa_monotone}).

Sollich \cite{Sollich2002b} analyzes the PA learning curve using the eigenvalue decomposition earlier covered. He assumes both student and teacher use kernels with the same eigenfunctions but possibly differing eigenvalues. %
Subsequently, he considers various synthetic distributions for which the eigenfunctions and eigenvalues can be computed analytically and 
finds that for a uniform distribution on the vertices of a hypercube, multiple overfitting maxima and plateaus may be present in the learning curve (see Figure \ref{fig:overview} \added{under (6.5)}), even if the student uses the teacher noise level.  
For a uniform distribution in one dimension, there may be arbitrarily many overfitting maxima if the student has a small enough noise level. In addition the convergence rates change and may become logarithmically slow.

The above analysis is extended by Sollich in \cite{Sollich2004}, where hyperparameters such as length scale and noise level, are now optimized during learning based on evidence maximization.
Among others, he finds that for the hypercube the arbitrary many overfitting maxima do not arise anymore and the learning curve becomes monotone.  %
All in all, Sollich concludes that optimizing the hyperparameters using evidence maximization can alleviate non-monotonicity.

\subsection{Misspecified Bayesian Regression} \label{sec:safebayes}

Gr\"{u}nwald and Van Ommen \cite{Grunwald2017} show that a (hierarchical) Bayesian linear regression model can give a broad peak in the learning curve of the squared risk, see Figure \ref{fig:overview} \added{under (6.6)}. %
This can happen if the homogeneous noise assumption is violated, while the estimator is otherwise consistent.

Specifically, let data be generated as follows. For each sample, a fair coin is flipped. Heads means the sample is generated according to the ground truth probabilistic model contained in the hypothesis class. Misspecification happens when the coin comes up tails and a sample is generated in a fixed location without noise. %

The peak in the learning curve cannot be explained by dipping, peaking or known sensitivity of the squared loss to outliers according to Gr\"{u}nwald and Van Ommen. The peak in the learning curve is fairly broad and occurs in various experiments. As also no approximations are to blame, the authors conclude that Bayes' rule is really at fault as it cannot handle the misspecification. %
The non-monotonicity can happen if the probabilistic model class is not convex. %

Following their analysis, a modified Bayes rule is introduced, in which the likelihood is raised to some power $\eta$.  The parameter $\eta$ cannot be learned in a Bayesian way, leading to their SafeBayes approach. Their technique alleviates the broad peak in the learning curve and is empirically shown to make the curves generally more well-behaved. %

\subsection{The Perfect Prior} \label{sec:perfect-prior}

As we have seen in Subsection \ref{sub:pa_monotone}, the PA learning curve is always monotone if the problem is well specified and a Bayesian decision theoretical approach is followed.  Nonetheless, the fact that the PA curve is monotone does not mean that the curve for every individual problem is. \cite{Grunwald2011} offers an insightful example (see also Figure \ref{fig:overview} \added{beneath (6.7)}):
consider a fair coin and let us estimate its probability $p$ of heads using Bayes' rule. We measure performance using the negative log-likelihood on an unseen coin flip and adopt a uniform Beta(1,1) prior on $p$. %
This prior, i.e., without any training samples, already achieves the optimal loss since it assigns the same probability to heads and tails. After a single flip, $n=1$, the posterior is updated and leads to a probabilities of $\tfrac{1}{3}$ or $\tfrac{2}{3}$ and the loss \emph{must} increase. Eventually, with $n \to \infty$, the optimal loss is recovered, forming a bump in the learning curve.  Note that this construction is rather versatile and can create non-monotonic behavior for practically any Bayesian estimation task.  In a similar way, any type of regularization can lead to comparable learning curve shapes (see also  \cite{Viering2019,loog2019minimizers} and Subsection \ref{sect:mono}).

An related example can be found in \cite{Al-Saleh2003}. It shows that the posterior variance can also increase for a single problem, unless the likelihood belongs to the exponential family and a conjugate prior is used. GPs  fall in this last class.

\subsection{Monotonicity: a General Fix?}

This section has noted a few particular approaches to restore monotonicity of a learning curve.  One may wonder, however, whether generally applicable approaches exist that can turn any learner into a monotone one.
A first attempt is made in \cite{Viering2020} which proposes a wrapper that, with high probability, makes any classifier monotone in terms of the the error rate. The main idea is to consider $n$ as a variable over which model selection is performed. When $n$ is increased, a model trained with more data is compared to the previously best model on validation data. Only if the new model is judged to be significantly better---following a hypothesis test, the older model is discarded. If the original learning algorithm is consistent and if the size of the validation data grows, the resulting algorithm is consistent as well. It is empirically observed that the monotone version may learn more slowly, giving rise to the question whether there always will be a trade-off between monotonicity and speed (refer to the learning curve in Figure \ref{fig:overview} \added{under (6.8)}). 

\cite{mhammedi2020risk} extended this idea, proposing two algorithms that do not need to set aside validation data while guaranteeing monotonicity. To this end they assume that the Rademacher complexity of the hypothesis class composited with the loss is finite. This allows them to determine when to switch to a model trained with more data. In contrast to \cite{Viering2020}, they argue that their second algorithm does not learn slower, as its generalization bound coincides with a known lower bound of regular supervised learning. 

\added{Recently, \cite{bousquet2022monotone} proved that, in terms of the 0-1 loss, any learner can be turned into a monotonic one. It extends a result obtained in \cite{pestov2022universally}, disproving a conjecture by Devroye et al. \cite[page 109]{Devroye1996} that universally consistent monotone classifiers do not exist.}

\section{Discussion and Conclusion}\label{sect:disc}

\added{Though there is some empirical evidence that for large deep learners, especially in the big data regime, learning curves behave like power laws, such conclusions seem premature.  For other learners results are mixed. Exponential shapes cannot be ruled out and there are various models that can perform on par with power laws.  Theory, on the other hand, often points in the direction of power laws and also supports exponential curves.}  GP learning curves have been analyzed for several special cases, but their general shape remains hard to characterize and offers rich behavior such as different regimes.  Ill-behaved learning curves illustrate that various shapes are possible that are hard to characterize. What is perhaps most surprising is that these behaviors can even occur in the well-specified case and realizable setting. It should be clear that, %
currently, there is no theory that covers all these different aspects.

\added{Roughly, our review sketches the following picture.} Much of the work on learning curves seems scattered and incidental.  Starting with Hughes, Foley, and Raudys, some initial contributions appeared around 1970. %
The most organized efforts, starting around 1990, come from the field of statistical physics, with important contributions from Amari, Tishby, Opper, and others. %
These efforts have found their continuation within GP regression, to which Opper again contributed significantly. For GPs, Sollich probably offered some of the most complete work. %

The usage of the learning curve as an tool for the analysis of learning algorithms has varied throughout the past decades.  In line with Langley, Perlich, Hoiem, et al.\ \cite{Langley1988,Perlich2003,hoiem2021learning}, we would like to suggest a more consistent use.  We specifically agree with Perlich et al.\ \cite{Perlich2003} that without a study of the learning curves, claims of superiority of one approach over the other are perhaps only valid for very particular sample sizes. Reporting learning curves in empirical studies can also help the field to move away from its fixation on bold numbers, besides accelerating learning curve research.

In the years to come, we expect investigations of parametric models and their performance in terms of extrapolation.  Insights into these problems become more and more important---particularly within the context of deep learning---to enable the intelligent use of computational resources. In the remainder, we highlight some specific aspects that we see as important. %

\subsection{Averaging Curves and The Ideal Parametric Model}

Especially for extrapolation, a learning curve should be predictable, which, in turn, asks for good parametric model. It seems impossible to find a generally applicable, parametric model that covers all aspects of the shape, in particular ill-behaving curves.  Nevertheless, we can try to find a  model class that would give us sufficient flexibility and extrapolative power. Power laws and exponentials should probably be part of that class, but does that suffice?  

To get close to the true learning curve, some studies average hundreds or thousands of individual learning curves \cite{Figueroa2012,Ng2002}.  Averaging can mask interesting characteristics of individual curves  \cite{schiavo2000ten,hand1997construction}. This has been extensively debated in psychonomics, where cases have been made for exponentially shaped individual curves, but power-law-like average curves \cite{anderson1997artifactual,heathcote2000power}. 
In applications, we may need to be able to model individual, single-training-set curves or curves that are formed on the basis of relatively small samples. As such, %
we see potential in studying and fitting individual curves to better understanding their behavior.%

\subsection{How to Robustly Fit Learning Curves}

A technical problem that has received little attention (except in \cite{hoiem2021learning,lcdb2022}) is how to properly fit a learning curve model.  As far as current studies at all mention how the fitting is carried out, they often seem to rely on simple least squares fitting of log values, assuming independent Gaussian noise at every $n$.  Given that this noise model is not bounded---while a typical loss cannot become negative, this choice seems disputable. At the very least, derived confidence intervals and $p$-values should be taken with a grain of salt.

\added{In fact, fitting can often fail, but this is not always reported in literature. In one of the few studies, \cite{lcdb2022}, where this is at all mentioned, 2\% of the fits were discarded. Therefore, further investigations into how to robustly fit learning curve models from small amounts of curve data seems promising to investigate.}  
In addition, a probabilistic model with assumptions that more closely match the intended use of the learning curve seems also worthwhile.  

\subsection{Bounds and Alternative Statistics} \label{sec_PAC}

One should be careful in interpreting theoretical results when it comes to the shape of learning curves. Generalization bounds, such as those provided by PAC, that hold uniformly over all $P$ may not correctly characterize the curve shape. Similarly, a strictly decreasing bounds does not imply monotone learning curves, and thus does not rule out ill-behavior. \added{We merely know that the curve lies under the bound, but this leaves room for strange behavior.} Furthermore, PA learning curves, which require an additional average over problems, can show behavior substantially different from those for a single problem, because here averaging can also mask characteristics.

Another incompatibility between typical generalization bounds and learning curves is that the former are constructed to hold with \emph{high probability} with respect to the sampling of the training set, while the latter look at the \emph{mean} performance over all training sets. Though bounds of one type can be transformed into the other \cite{mey2020note}, this conversion can change the actually shape of the bound, thus such high probability bounds may also not correctly characterize learning curve shape for this reason.

The preceding can also be a motivation to actually study learning curves for statistics other than the average. For example, in Equation \ref{eq_exp_lc}, instead of the expectation we could look at the curves of the median or other percentiles. These quantities are closer related to high probability learning bounds.  Of course, we would not have to choose the one learning curve over the other.  They offer different types of information and, depending on our goal, may be worthwhile to study next to each other.  Along the same line of thought, we could include quartiles in our plots, rather than the common error bars based on the standard deviation.  Ultimately, we could even try to visualize the full loss distribution at every sample size $n$ and, potentially, uncover behavior much more rich and unexpected.

A final estimate that we think should be investigates more extensively is the training loss.  Not only can this quantity aid in identifying overfitting and underfitting issues \cite{hand1997construction,jain2000statistical}, but it is a quantity that is interesting to study as such or, say, in combination with the true risk.  Their difference, a simple measure of overfitting, could, for example, turn out to behave more regular than the two individual measures.  

\subsection{Research into Ill-Behavior and Meta-learning}

We believe better understanding is needed regarding the occurrence of peaking, dipping, and otherwise non-monotonic or phase-transition-like behavior: when and why does this happen? Certainly, a sufficiently valid reason to investigate these phenomena is to quench one's scientific curiosity.  We should also be careful, however, not to mindlessly dismiss such behavior as mere oddity.  Granted, these uncommon and unexpected learning curves have often been demonstrated in artificial or unrealistically simple settings, but this is done to make at all insightful that there is a problem. 

The simple fact is that, at this point, we do not know what role these phenomena play in real-world problems. \added{\cite{lcdb2022} shared a database of learning curves for 20 learners on 246 datasets. These questions concerning curious curves can now be investigated in more detail, possibly even in automated fashion. } 
Given the success of meta-learning using learning curves this seems a promising possibility. Such meta-learning studies on large amounts of datasets could, in addition, shed more light on what determines the parameters of learning curve models, a topic that has been investigated relatively little up to now. Predicting these parameters robustly from very few points along the learning curve will prove valuable for virtually all applications.

\subsection{Open Theoretical Questions}\label{sect:theo}

There are two specific theoretical questions that we would like to ask.  Both are concerned with the monotonicity.

\added{One interesting question that remains is whether monotonic learners can be created for losses other than 0-1, unbounded ones in particular.  Specifically, we wonder whether maximum likelihood estimators for well-specified models behave monotonically.  Likelihood estimation, being a century-old, classical technique \cite{stigler2007epic}, has been heavily studied, both theoretically and empirically.  In much of the theory developed, the assumption that one is dealing with a correctly specified model is common, but we are not aware of any results that demonstrate that better models are obtained with more data.  The question is interesting because this estimator has been extensively studied already and still plays a central role in statistics and abutting fields.}

\added{Secondly, as ERM remains one of the
primary inductive principles in machine learning, another question that remains interesting---and which was raised in \cite{loog2019minimizers,bousquet2022monotone} as well, is when  ERM \emph{will} result in monotonic curves?}

\subsection{Concluding}

More than a century of learning curve research has brought us quite some insightful and surprising results.  What is  more striking however, at least to us, is that there is still so much that we actually do not understand about learning curve behavior.  Most theoretical results are restricted to relatively basic learners, while much of the empirical research that has been carried out is quite limited in scope.  In the foregoing, we identified some specific challenges already, but we are convinced that many more open and interesting problems can be discovered.  In this, the current review should help both as a guide and as a reference.

\subsection*{Acknowledgments}

We would like to thank Peter Gr\"{u}nwald, Alexander Mey, Jesse Krijthe, Robert-Jan Bruintjes, \added{Elmar Eisemann} \added{and three anonymous reviewers} for their valueable input.

\bibliographystyle{IEEEtran}
\bibliography{main}

\begin{IEEEbiography}[{\includegraphics[width=1in,height=1.25in,clip,keepaspectratio]{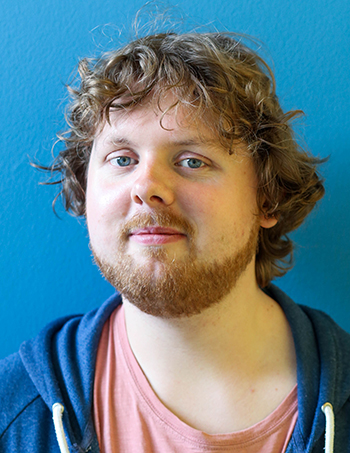}}]{Tom Viering} received the MSc degree in computer science from the Delft University of Technology in 2016 and is currently a PhD student in the Pattern Recognition Laboratory there. His principal research interests are statistical learning theory and its application to problems such as active learning, domain adaptation, and semi-supervised learning. Other topics that interest him are non-monotonicity of learning, interpretable machine learning, generative models, online learning and deep learning. 
\end{IEEEbiography}
\begin{IEEEbiography}[{\includegraphics[width=1in,height=1.25in,clip,keepaspectratio]{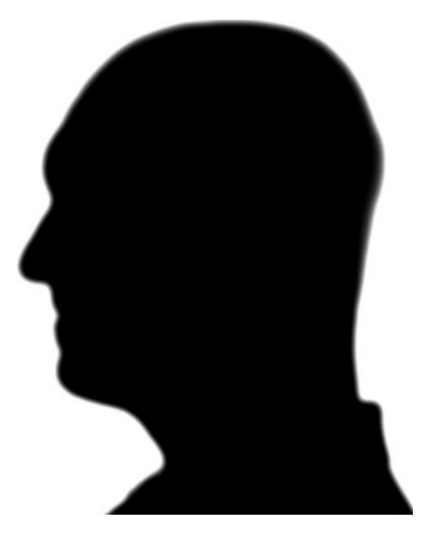}}]{Marco Loog} received an MSc degree in mathematics from Utrecht University and a PhD degree from the Image Sciences Institute. He worked as a scientist at the IT University of Copenhagen, the University of Copenhagen, and Nordic Bioscience. He now is at Delft University of Technology to research and to teach. Marco is an honorary professor at the University of Copenhagen. His principal research interest is with supervised learning in all sorts of shapes and sizes.
\end{IEEEbiography}


\end{document}